\documentclass{IEEEtran}
\usepackage[caption=false]{subfig}
\usepackage[utf8]{inputenc}
\usepackage[bottom]{footmisc}
\usepackage[boxed,lined,linesnumbered]{algorithm2e}
\usepackage[colorlinks=true, allcolors=blue]{hyperref}
\usepackage{times, amsfonts, amssymb, amsmath, textcomp, gensymb, epsfig, makeidx, graphicx, graphics, url, multirow, multicol, booktabs, array, bm, rotating, float, setspace, paralist, wrapfig, xcolor, enumitem, latexsym, longtable, tabularx, xcolor, cite, mathtools, stmaryrd, commath, colortbl, pgfplotstable, tikz, collcell}
\usepackage[slantedGreek]{mathptmx}
\graphicspath{{./Figures/}}
\DeclareMathAlphabet{\mathcal}{OMS}{cmsy}{m}{n}

\makeatletter
\renewcommand*\env@matrix[1][c]{\hskip -\arraycolsep
  \let\@ifnextchar\new@ifnextchar
  \array{*\c@MaxMatrixCols #1}}
\makeatother

\newcommand{\etal}{\textit{et al}.,}

\begin{document}

\title{An Instance Space Analysis of Constrained Multi-Objective Optimization Problems}

\author{Hanan~Alsouly,
        Michael~Kirley,
        and~Mario~Andrés~Mu\~{n}oz,
\thanks{Funding was provided by the Australian Research Council through grants FL140100012 and IC200100009.}
\thanks{H. Alsouly, M. Kirley and M.A. Mu\~{n}oz are with the School of Computing and Information, University of Melbourne, Melbourne, Australia, and with the ARC Centre in Optimisation Technologies, Integrated Methodologies and Applications (OPTIMA),  Australia. e-mail: (halsouly@student.unimelb.edu.au and \{mkirley,munoz.m\}@unimelb.edu.au).}
\thanks{H. Alsouly is also with  College of Computing and Information Sciences, Imam Mohammad Ibn Saud Islamic University (IMSIU), Riyadh, Saudi Arabia}
\thanks{Manuscript received MMM D, 2022; revised MMM D, 2022.}}

%



\maketitle

\begin{abstract}

Multi-objective optimization problems with constraints (CMOPs) are generally considered more challenging than those without constraints. This in part can be attributed to the creation of infeasible regions generated by the constraint functions, and/or the interaction between constraints and objectives. In this paper, we explore the relationship between constrained multi-objective evolutionary algorithms (CMOEAs) performance and CMOP instances characteristics using Instance Space Analysis (ISA). To do this, we extend recent work focused on the use of Landscape Analysis features to characterise CMOP. Specifically, we scrutinise the multi-objective landscape and introduce new features to describe the multi-objective-violation landscape, formed by the interaction between constraint violation and multi-objective fitness. Detailed evaluation of problem-algorithm footprints spanning six CMOP benchmark suites and fifteen CMOEAs, illustrates that ISA can effectively capture the strength and weakness of the CMOEAs. We conclude that two key characteristics, the isolation of non-dominate set and the correlation between constraints and objectives evolvability, have the greatest impact on algorithm performance. However, the current benchmarks problems do not provide enough diversity to fully reveal the efficacy of CMOEAs evaluated.
\end{abstract}


%
\IEEEpeerreviewmaketitle

\section{Introduction}\label{sec:introduction}


\IEEEPARstart{C}{onstrained} multi-objective optimization problems (CMOPs) involve searching for the best trade-off between multiple conflicting objectives subject to one or more constraints. Many real-world optimization problems match this description, in areas as diverse as mechanical design, chemical engineering, and power system optimization~\cite{Abhishek2021}. Generally, a CMOP is more challenging than its unconstrained counterpart due to the addition of one or more constraint functions, and the resulting interactions between the constraints and the objectives ~\cite{Fan2020}. Constraints may change the shape and location of the Pareto front $\left(PF\right)$, often creating only small feasible regions, resulting in additional difficulties when attempting to generate a diverse set of solutions on the $PF$. Consequently, several constrained multi-objective evolutionary algorithms (CMOEAs) have been introduced to specifically tackle CMOPs~\cite{Bogdan2021}. However, as per many other problem domains, practice has shown that no single algorithm outperforms all other algorithms across all problem instances~\cite{Fan2017, Abhishek2021}. Each algorithm has its strengths and weaknesses, and it is hard to choose the best one for solving a particular instance. Therefore, it is necessary to understand when an algorithm is suitable or not, i.e., when it performs well and when it fails, which requires an understanding of the characteristics of the instances being solved, e.g. multi-modality and variable scaling, and what distinguish the instances from each other~\cite{Mersmann2011}.

\begin{figure}[!t]
	\centering
	\includegraphics[width=0.48\textwidth, clip]{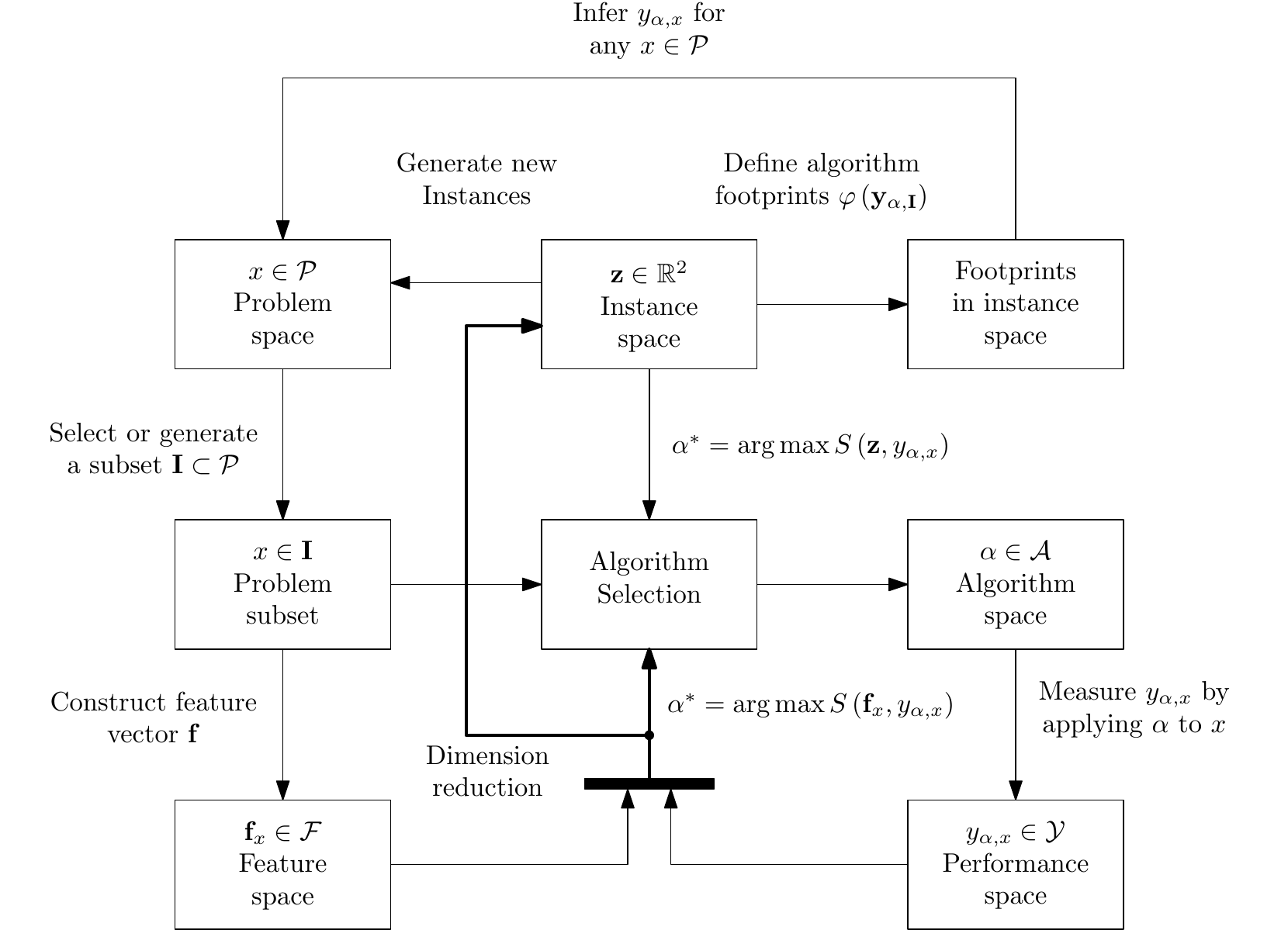}
	\caption{Summary of the Instance Space Analysis framework~\protect\cite{SmithMiles2021}.}
	\label{fig:AlgorithmSelection}
\end{figure}

In this paper, we explore the relationship between algorithm performance and CMOP instances characteristics using Instance Space Analysis (ISA). Proposed by Smith-Miles \etal~\cite{SMITHMILES2014}, ISA is a methodology for assessing the difficulty of a set of problem instances for a group of algorithms. Figure~\ref{fig:AlgorithmSelection} illustrates ISA's framework, which uses a meta-data set consisting of features that characterize a set of instances, and performance measures of a group of algorithms on those instances. Then, by selecting a subset of uncorrelated features that are predictive of algorithm performance, and using a tailored dimensionality reduction method, the meta-data is projected into a 2-dimensional plane called the \textit{instance space}. In it, each instance is represented as a point, allowing for the visualization of the similarities and differences between instances, in terms of characteristics and algorithm performance. An examination of the generated instance space can then be used to identify regions of good performance, called \textit{footprints}, where an algorithm is expected to perform well and why.

ISA has been employed successfully on related problem domains. For example, Yap \etal~\cite{Yap2020} performed an ISA of combinatorial multi-objective optimization problems (MOPs), discovering that MOEA/D is preferred, not only when the number of objectives increased, but also when the degree of conflict between objectives decreased. Similarly, Mu\~{n}oz and Smith-Miles~\cite{Muoz2017PerformanceAO} analyzed the space of continuous single-objective optimization problems, identifying that multi-modal instances with adequate global structure are hard to solve by most studied algorithms with exception to BIPOP-CMA-ES. In both works, Landscape Analysis \textit{features}~\cite{Malan2021} were employed to characterize a problem instance. Therefore, a necessary first step when applying ISA for CMOP will be to identify and calculate appropriate, informative Landscape Analysis features. 

Recent work have proposed Landscape Analysis features for characterizing MOPs. Kerschke \etal~\cite{Kerschke2018} studied the notion of multimodality in MOPs and provided a set of features to quantify it, whilst Liefooghe \etal~\cite{Liefooghe2021} extended previous works in the combinatorial MOPs domain to characterize continuous MOPs, focusing on multimodality, evolvability, and ruggedness. Unfortunately, it is not a straight forward task to identify Landscape Analysis features for CMOP. The features described above can be used in the CMOP domain to help characterize the objective space. However, features associated with the constraints violation, and features representing the interaction between objectives and constraints are required. 

There have been only a few attempts to characterize constrained optimization problems. For example, for single-objective problems Malan \etal~\cite{Malan2015} defined the concept of a Violation landscape, proposing four features to characterize the feasible and constrained spaces. In other work, Poursoltan and Neumann~\cite{Poursoltan2015} introduced a biased sampling technique to quantify the ruggedness of a constrained optimization problems. Picard and Schiffmann~\cite{Picard2020} focussing on CMOPs, adopted two features from~\cite{Malan2015}, and extended another so that it could be used to measure `disjointedness' of the feasible regions. They also proposed two features to quantify the relationship between objectives and constraints. Vodopija \etal~\cite{Vodopija2021} were the first to introduce violation multimodality in CMOPs, proposing a set of features to characterize violation multimodality, smoothness, and the correlation between the objectives and constraints. They then used those features to compare the characteristics of eight benchmark problem suites against a real-world suite. This work did provide important insights, however, the approach was limited to the Violation landscape and did not capture important aspects that need to be quantified, such as the relationship between the constrained and unconstrained PF, or the ruggedness and evolvability of the multi-objective-violation landscape.

To construct the first ISA for continuous CMOP, we introduce the multi-objective-violation landscape, formed by the interaction between constraint violation and multi-objective fitness. This required the introduction of 12 new features and modification of 22 existing features to quantify the characteristics of the Violation landscape and multi-objective-violation landscape. The meta-data set was then generated for the instance space by processing the features for six CMOP benchmark suites and the performance of 15 CMOEAs. 

Comprehensive analysis of the generated footprints illustrates that ISA can effectively capture the strength and weakness of the CMOEAs. A key observation is that there are two characteristics in particular that affect the performance of most CMOEAs -- the isolation of the non-dominate set and the correlation between constraints and objectives. However, the performance of each CMOEA is affected by a different set of features. Importantly, the footprints provided strong supporting visual evidence as to which characteristics are necessary if any new proposed benchmarks are to significantly challenge CMOEAs.

The remainder of the paper is organized as follows: Section \ref{sec:ISA} presents the ISA methodology and its components, which are the problem include CMOPs definition, benchmark suites, landscape features, CMOEAs, and performance metric. Section \ref{sec:Exp} describes the experimental setup. The results are presented and discussed in section \ref{sec:results}. Finally, section \ref{sec:conclusion} concludes the paper.

\section{Instance Space Analysis}\label{sec:ISA}

Instance Space Analysis (ISA) traces its foundations to Rice's framework for solving the \textit{Algorithm Selection Problem}~\cite{Rice1976}, which suggests the construction of a selection mapping between measurable features of a problem and a set of suitable algorithms; and Wolpert and Macready's \textit{No-Free Lunch theorems}~\cite{Wolpert1997}, which state that an algorithm is unlikely to outperform all other algorithms on all possible instances. Figure~\ref{fig:AlgorithmSelection} illustrates ISA's framework, which has at its core six component \textit{spaces} or sets~\cite{SmithMiles2021}:
\begin{inparaenum}[(a)]
	\item the \textit{problem space}, $\mathcal{P}$, containing all the relevant instances of a problem in an application domain;
	\item a \textit{subset of instances}, $\mathbf{I}$, for which we have meta-data from computational experiments;
	\item the \textit{feature space}, $\mathcal{F}$, which are used to characterize the mathematical and statistical properties of the instances;
	\item the \textit{algorithm space}, $\mathcal{A}$, representing the set of algorithms available to solve all instances in $\mathbf{I}$;
	\item the \textit{performance space}, $\mathcal{Y}$, composed of a measure of the computational effort to obtain a satisfactory solution; and
	\item the \textit{instance space}, a 2-dimensional visualization that aids on the observation of trends in hardness for different algorithms, and facilitates insights into the distribution of existing instances.
\end{inparaenum}

This section describes the details for each one of the spaces in the ISA framework, tailored specifically for CMOPs. We start with $\mathcal{P}$ by formally defining a CMOP. Then, we present $\mathbf{I}$, drawn from six commonly used benchmark suites. Next, we describe $\mathcal{F}$, where we introduce the multi-objective-violation Landscape, as well as a set of new features for characterizing CMOPs. We follow by describing $\mathcal{A}$, by briefly presenting the 15 algorithms under test, and $\mathcal{Y}$ by formally defining the \textit{hyper-volume}, our performance metric.

\subsection{Problem Space}\label{subsec:CMOP}

\begin{figure*}[!t]
	\centering
	\subfloat[][]{
		\includegraphics[width=0.24\textwidth]{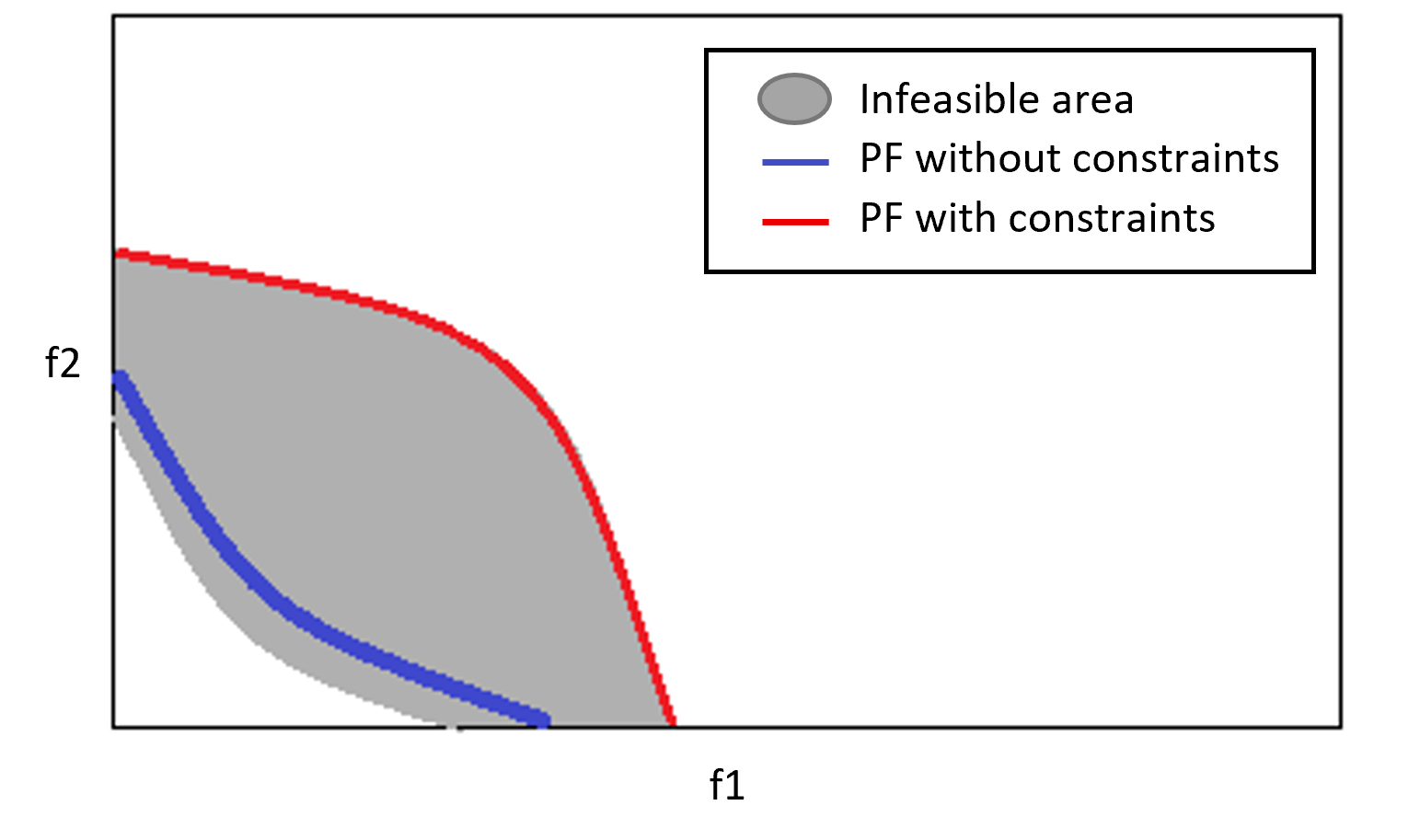}
		\label{fig:CPF1}
	}%
	\subfloat[][]{
		\includegraphics[width=0.24\textwidth]{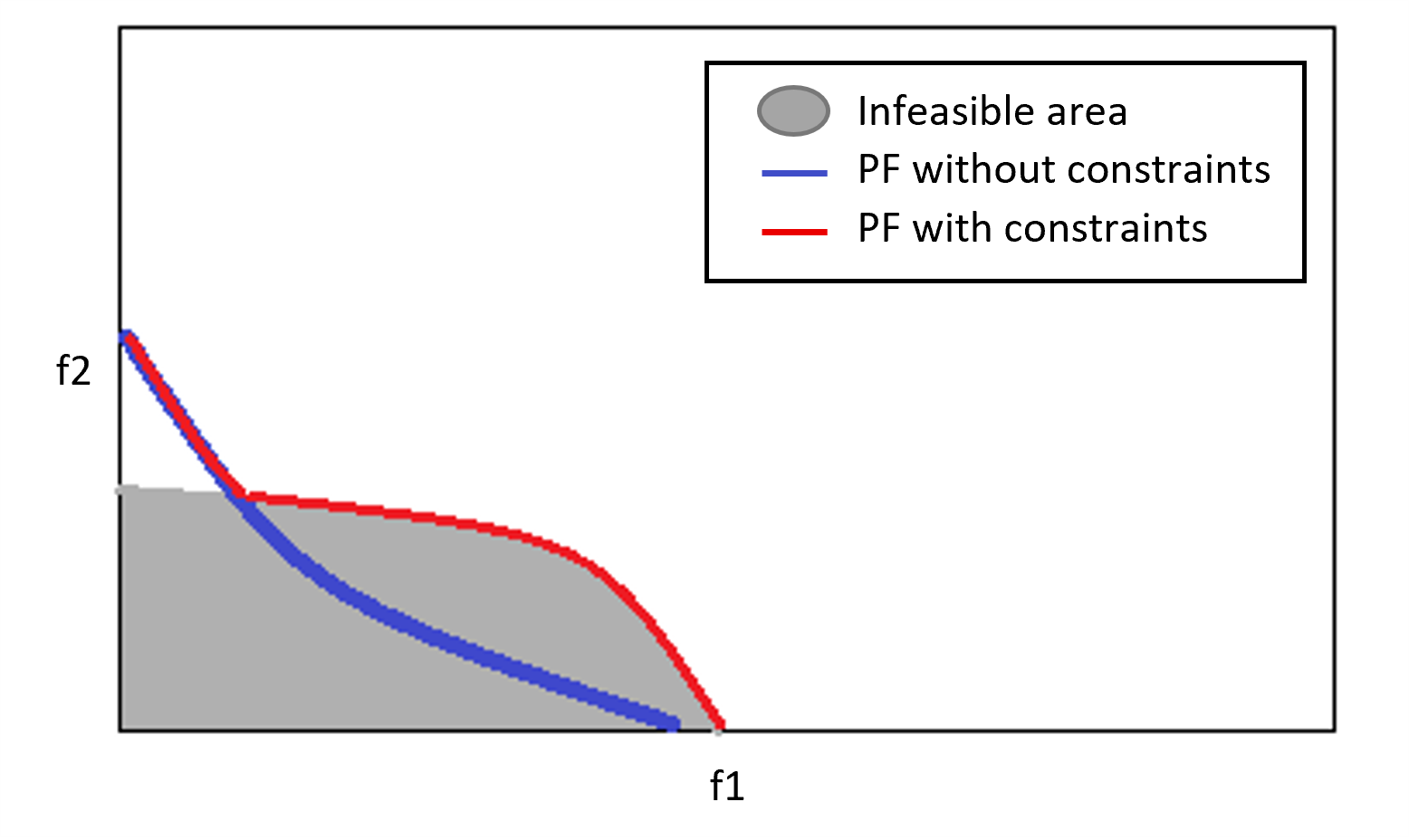}
		\label{fig:CPF2}
	}%
	\subfloat[][]{
		\includegraphics[width=0.24\textwidth]{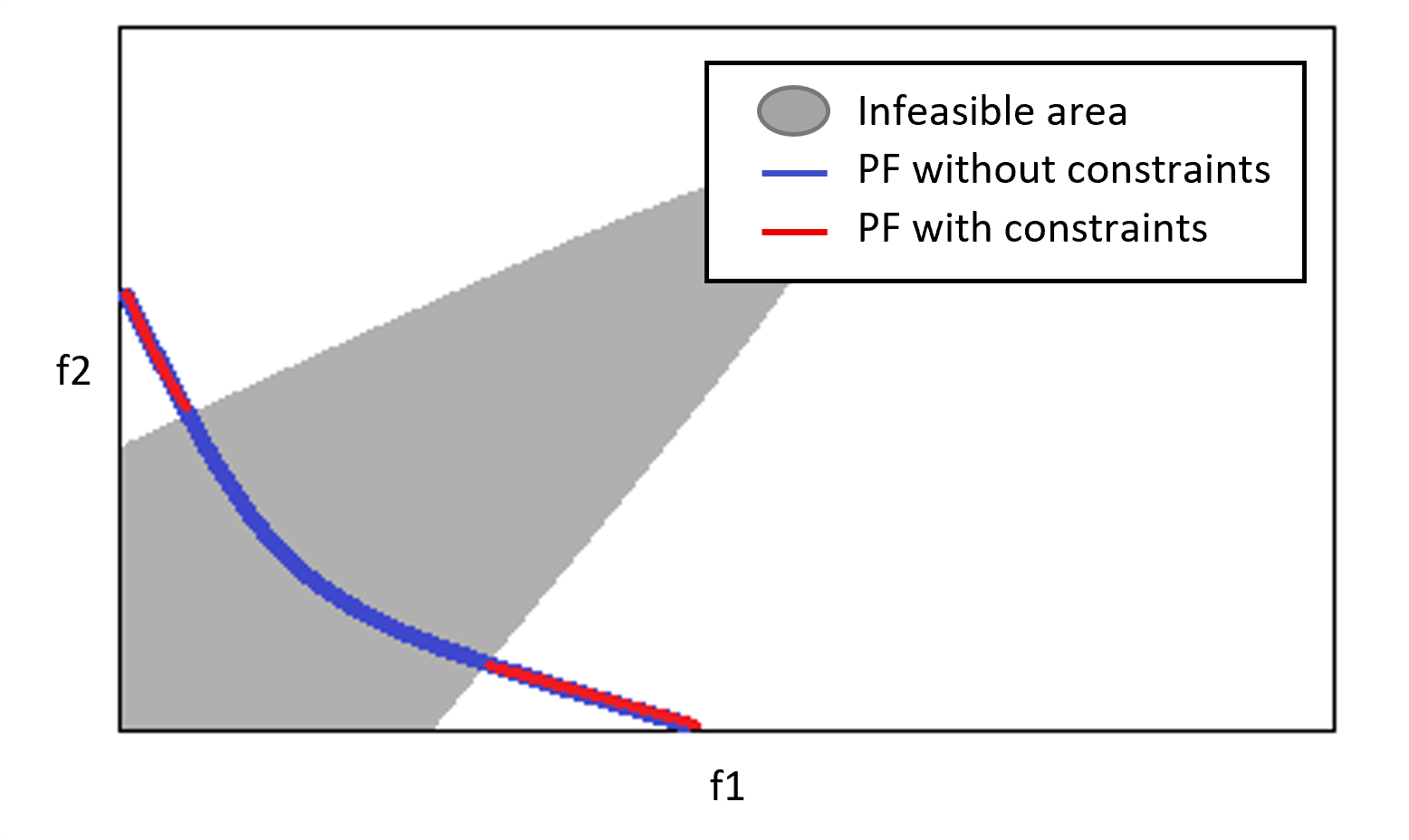}
		\label{fig:CPF3}
	}%
	\subfloat[][]{
		\includegraphics[width=0.24\textwidth]{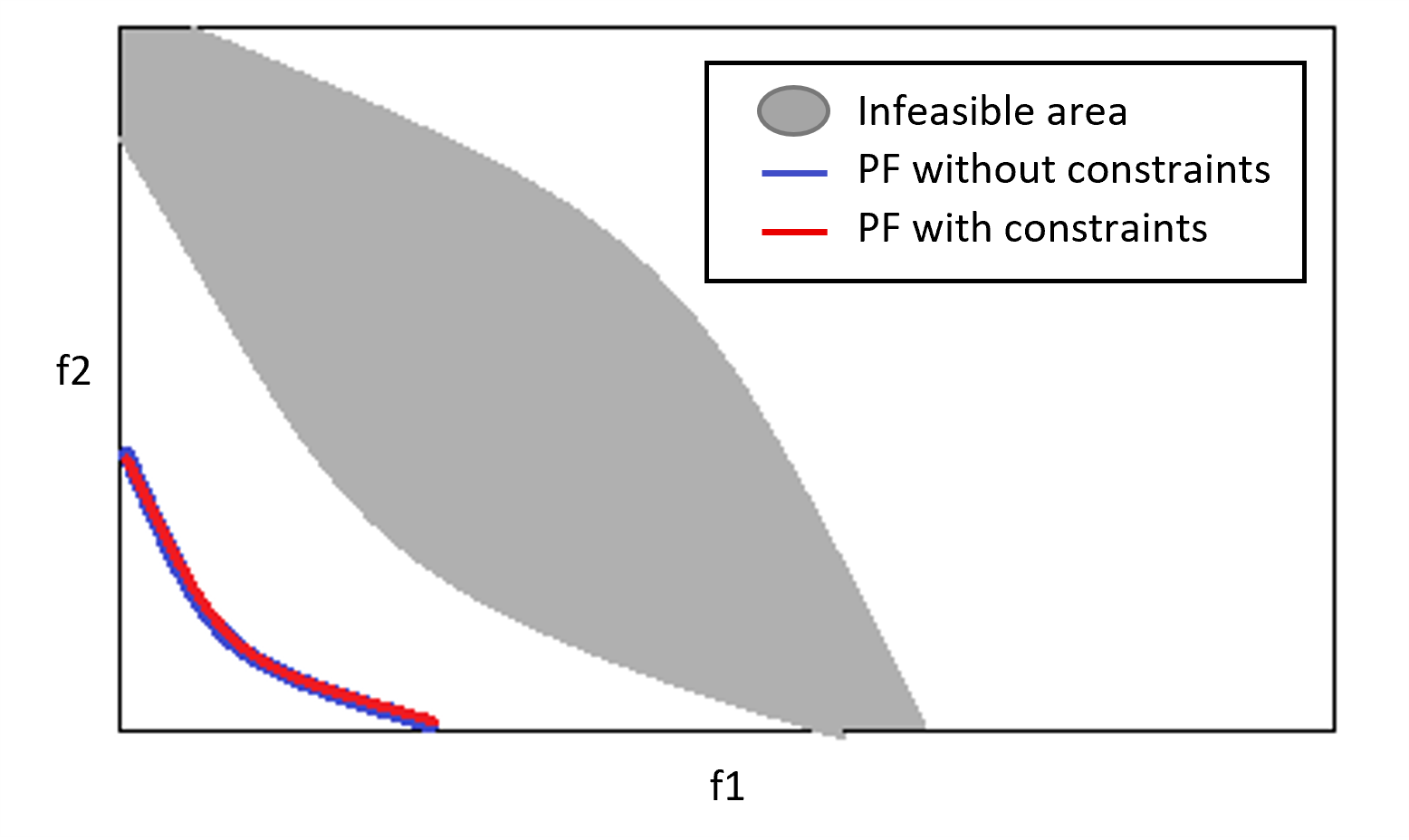}
		\label{fig:CPF4}
	}
	\caption{Effect of constraints on the $PF$: \protect\subref{fig:CPF1} The $UPF$ is no longer feasible, and the true $PF$ lies completely on bounds of the feasible region; \protect\subref{fig:CPF2}  part of the $UPF$ is no longer feasible, and parts of the true $PF$ lies on bounds of the feasible region; \protect\subref{fig:CPF3} the true $PF$ is only a proportion of the $UPF$; and \protect\subref{fig:CPF4} an infeasible region blocks the way toward the $PF$.}
	\label{fig:PF}
\end{figure*}

A CMOP can be defined as finding a vector of decision variables that optimizes a set of objective functions and satisfies a set of restrictions. Without loosing generality, we assume minimization. CMOP can be mathematically defined as follows:
\begin{equation}
	\left.
	\begin{array}{rll}
		\min & f_{m}(x), & m=1,2, \ldots, M \\
		\text{s.t.} & g_{j}(x) \geq 0, & j=1,2, \ldots, J \\
		& h_{k}(x)=0, & k=1,2, \ldots, K \\
		& x_{i}^{(L)} \leq x_{i} \leq x_{i}^{(U)}, & i=1,2, \ldots, n
	\end{array}
	\right\}
	\label{eq:cmop}
\end{equation}
\noindent where a solution $x\in R^n$ is a vector of $n$ decision variables, $f(x)$ is a vector of $M$ objectives to be optimized, $g(x)$ is set of $J$ inequality constraints, while $h(x)$ is set of $K$ equality constraints, and $x_{i}^{(L)}$ and $x_{i}^{(U)}$ represent respectively the lower and upper bounds of a decision variable $x_i$. The constraints violation of a solution $x$ can be calculated by the following equation:
\begin{equation}
	CV\left({x}\right)=\sqrt{
		\sum_{j=1}^{J} G_{j}\left({x}\right)^{2}
		+ \sum_{k=1}^{K} H_{k}\left({x}\right)^{2}}
	\label{eq:CV}
\end{equation}
\noindent where 
\begin{equation}
	G_{j}\left({x}\right)=\max \left(0, g_{j}\left({x}\right)\right)
\end{equation}
\noindent and
\begin{equation}
	H_{k}\left({x}\right)=\max \left(0,\left|h_{k}\left({x}\right)\right|-\epsilon\right)
\end{equation}
\noindent where $\epsilon$ is a small value $\left({10}^{-4}\right)$ to relax the equality constraints. A solution to the problem is \textit{feasible} when it satisfies the sets of constraints, otherwise the solution is \textit{infeasible}.

A solution $x\in R^n$ is said to be \textit{Pareto optimal} if there is no other feasible solution $y\in R^n$ such that $f(y)\prec f(x)$, where $\prec$ indicates the Pareto dominance relation. That is, a feasible solution $x$ dominates a feasible solution $y$ if and only if no objective value of $f(x)$ is greater than $f(y)$ and at least one objective value is smaller. Because of the conflicting nature of the multiple objectives, optimizing one objective function may lead to a degradation in another. Therefore, a single optimal solution may no longer be found, but instead a set of trade-off solutions, the so-called non-dominated solutions or Pareto optimal set ($PS$). Those set of solutions represent the Pareto optimal front ($PF$) in the objective space.

When solving a MOP, an algorithm aims to find an estimated front, $\widetilde{PF}$, that has \textit{converged}, i.e., is as close as possible to the $PF$, and is \textit{diverse}, i.e., represents the whole $PF$. In CMOPs, the feasibility of the solution(s) must also be considered. The true, constrained, $PF$ of a CMOP can be determined by the unconstrained $PF$ ($UPF$) and bounds of the feasible regions in the objective space. Having a low proportion of feasible regions typically adds to the challenge/difficulty of the search process. In addition, the infeasible regions may affect the shape of the $PF$ or split it into many segments, which may impact the algorithm’s ability to provide diversity in its solutions. That is, the infeasible regions may block the trajectory of the search towards $PF$, limiting convergence~\cite{Fan2020}. Figure~\ref{fig:PF} illustrates examples of difficulties caused by constraints.

\subsection{Subset of instances}\label{subsec:bench}

Test problems should cover as many characteristics of real-world problems as possible, if an evaluation of the performance of the optimization algorithm is to provide useful insights before solving an expensive real-world problem. Several CMOP benchmarks have been designed with this goal in mind. Ma and Wang~\cite{Zhongwei2019} proposed a classification of CMOPs depending on the relationship between $UPF$ and the $PF$:

\begin{description}
	\item[Type I] the $PF$ is same as the $UPF$.
	\item[Type II] the $PF$ is part of the $UPF$.
	\item[Type III] the $PF$ contains all or part of the $UPF$ and solutions on the boundary of the feasible region.
	\item[Type IV] the $PF$ is completely located on the boundary of the feasible region.
\end{description}

We have used six benchmark suites in this work: CF~\cite{CEC2009}, C-DTLZ~\cite{Jian2014}, DC-DTLZ~\cite{Chen2018}, LIR-CMOP~\cite{Fan2019}, DAS-CMOP~\cite{Fan2020}, and MWs~\cite{Zhongwei2019}. These suites have a wide range of characteristics, which are summarized in Table~\ref{tab:bench}.

\begin{table}[!t]
	\caption{Characteristics of the benchmarks employed in this study. All of them have a scalable number of decision variables. \textit{Type} describes the relationship between $UPF$ and $PF$, $M$ is the number of objectives, $CF$ is the number of constraints, $UPF$ and $PF$ columns define their shape, and the size and connectivity of the feasible region is given in the last column.  \textit{S} is short for scalable, \textit{C} for Controllable, \textit{Disconn} for Disconnected, \textit{Conn} for Connected, \textit{L} for Large, and \textit{S} for Small.}
	\label{tab:bench}
	\scriptsize
	\begin{tabular}{llllllll}
		\toprule
		\textbf{Problem} & \textbf{Type} & \textbf{M} & \textbf{CF} & \textbf{UPF} & \textbf{PF} & \multicolumn{2}{c}{\textbf{Feasible   Region}} \\ \midrule
		CF1              & II            & 2          & 1          & Linear       & Disconn & L/Conn                       \\
		CF2              & II            & 2          & 1          & Convex       & Disconn & L/Conn                       \\
		CF3              & II            & 2          & 1          & Concave      & Disconn & L/Conn                       \\
		CF4              & III           & 2          & 1          & Linear       & Linear       & L/Conn                      \\
		CF5              & III           & 2          & 1          & Linear       & Linear       & L/Conn                      \\
		CF6              & III           & 2          & 1          & Convex       & Mixed        & L/Conn                       \\
		CF7              & III           & 2          & 1          & Convex       & Mixed        & L/Conn                       \\
		CF8              & II            & 3          & 1          & Concave      & Disconn & L/Conn                       \\
		CF9              & II            & 3          & 1          & Concave      & Disconn & L/Conn                       \\
		CF10             & II            & 3          & 1          & Concave      & Disconn & L/Conn                       \\
		C1-DTLZ1         & I             & S          & 1          & Linear       & Linear       & S/Conn                       \\
		C2-DTLZ2         & II            & S          & 1          & Concave      & Concave      & S/Disconn                    \\
		C1-DTLZ3         & I             & S          & 1          & Concave      & Concave      & L/Disconn                    \\
		C3-DTLZ4         & IV            & S          & M          & Concave      & Concave      & L/Conn                       \\
		DC1-DTLZ1        & II            & S          & 1          & Linear       & Disconn & C/Disconn                    \\
		DC1-DTLZ3        & II            & S          & 1          & Concave      & Disconn & C/Disconn                    \\
		DC2-DTLZ1        & I             & S          & 2          & Linear       & Linear       & S/Disconn                    \\
		DC2-DTLZ3        & I             & S          & 2          & Concave      & Concave      & S/Disconn                    \\
		DC3-DTLZ1        & II            & S          & M          & Linear       & Disconn & C/Disconn                    \\
		DC3-DTLZ3        & II            & S          & M          & Concave      & Disconn & C/Disconn                    \\
		DAS-CMOP1        & C             & 2          & 11         & Concave      & Disconn & C/C                    \\
		DAS-CMOP2        & C             & 2          & 11         & Mixed        & Mixed        & C/C                    \\
		DAS-CMOP3        & C             & 2          & 11         & Disconn & Disconn & C/C                    \\
		DAS-CMOP4        & C             & 2          & 11         & Concave      & Disconn & C/C                    \\
		DAS-CMOP5        & C             & 2          & 11         & Mixed        & Mixed        & C/C                    \\
		DAS-CMOP6        & C             & 2          & 11         & Disconn & Disconn & C/C                    \\
		DAS-CMOP7        & C             & 3          & 7          & Linear       & Disconn & C/C                    \\
		DAS-CMOP8        & C             & 3          & 7          & Concave      & Disconn & C/C                    \\
		DAS-CMOP9        & C             & 3          & 7          & Concave      & Disconn & C/C                    \\
		LIR-CMOP1         & IV            & 2          & 2          & Concave      & Concave      & S/Conn                       \\
		LIR-CMOP2         & IV            & 2          & 2          & Convex       & Convex       & S/Conn                       \\
		LIR-CMOP3         & IV            & 2          & 3          & Concave      & Disconn & S/Disconn                    \\
		LIR-CMOP4         & IV            & 2          & 3          & Convex       & Disconn & S/Disconn                    \\
		LIR-CMOP5         & I             & 2          & 2          & Convex       & Convex       & S/Disconn                    \\
		LIR-CMOP6         & I             & 2          & 2          & Concave      & Concave      & S/Disconn                    \\
		LIR-CMOP7         & IV            & 2          & 3          & Convex       & Concave      & S/Disconn                    \\
		LIR-CMOP8         & IV            & 2          & 3          & Concave      & Concave      & S/Disconn                    \\
		LIR-CMOP9         & II            & 2          & 2          & Concave      & Disconn & S/Disconn                    \\
		LIR-CMOP10        & II            & 2          & 2          & Convex       & Disconn & S/Disconn                    \\
		LIR-CMOP11        & III           & 2          & 2          & Convex       & Disconn & S/Disconn                    \\
		LIR-CMOP12        & III           & 2          & 2          & Concave      & Disconn & S/Disconn                    \\
		LIR-CMOP13        & I             & 3          & 2          & Mixed        & Mixed        & S/Disconn                    \\
		LIR-CMOP14        & II            & 3          & 3          & Mixed        & Mixed        & S/Disconn                    \\
		MW1              & II            & 2          & 1          & Linear       & Disconn & S/Disconn                    \\
		MW2              & I             & 2          & 1          & Linear       & Linear       & S/Disconn                    \\
		MW3              & III           & 2          & 2          & Linear       & Mixed        & S/Conn                       \\
		MW4              & I             & S          & 1          & Linear       & Linear       & S/Conn                       \\
		MW5              & II            & 2          & 3          & Concave      & Disconn & S/Conn                       \\
		MW6              & II            & 2          & 1          & Concave      & Disconn & S/Disconn                   \\
		MW7              & III           & 2          & 2          & Concave      & Disconn & S/Conn                       \\
		MW8              & II            & S          & 1          & Concave      & Disconn & S/Disconn                    \\
		MW9              & IV            & 2          & 1          & Convex       & Concave      & S/Conn                       \\
		MW10             & III           & 2          & 3          & Concave      & Disconn & S/Disconn                    \\
		MW11             & IV            & 2          & 4          & Concave      & Disconn & S/Disconn                    \\
		MW12             & IV            & 2          & 2          & Mixed        & Mixed        & S/Disconn                    \\
		MW13             & III           & 2          & 2          & Disconn & Disconn & S/Disconn                    \\
		MW14             & I             & S          & 1          & Disconn & Disconn & S/Conn                       \\ \bottomrule
	\end{tabular}
\end{table}

\subsection{Feature Space}\label{subsec:CMOLandscapes}

Landscape Analysis are methods used to quantify the characteristics of a problem's landscape, which is described as a surface in the search space that defines a certain aspect of the problem, such as fitness, for each potential solution~\cite{Mersmann2011}. Stadler~\cite{Stadler2002} defined a general form of the fitness landscape for a problem as the triplet $\left(X, N, f\right)$, where $X$ is a set of potential solutions, $f$ is a fitness function, and $N$ is a notion of neighborhood relation. The Euclidean distance is usually used in continuous optimization to quantify solutions relations.

Constrained multi-objective optimization involves multiple fitness and constraint functions; hence, Stadler's definition of fitness landscape cannot be directly applied in this work. Verel \etal~\cite{Verel2011} defined the multi-Objective landscape, and Malan \etal~\cite{Malan2015} introduced the violation landscape. However, these two landscapes treat constraints and objectives independently. Therefore, they do not capture the interaction between them, which is essential for CMOPs. Consequently, we introduce the concept of the multi-objective-violation landscape in~\ref{subsec:MOCVLndscape}. Also, we propose a set of local structured-based landscape features collected by random walks, and global unstructured-based landscape features approximated by random samples~\cite{Munoz2015}. In the next subsections, we present the three landscapes and their features in detail.

\begin{table*}[!t]
	\centering
	\caption{The features used to characterize the Multi-Objectives Landscape of CMOP.}
	\label{tab:features3}
	\scriptsize
	\begin{tabular}{llp{9cm}lp{2cm}}
		\toprule
		Type                         & \textbf{Feature}           & \textbf{Description}                                                                                    & \textbf{Source}      & \textbf{Focus}   \\ \midrule
		\multirow{18}{*}{Global}     & upo\_n                     & Proportion of unconstrained PO solutions                                                                & \cite{Knowles2003}   & Set-Cardinality  \\
		                             & uhv                        & Hypervolume-value of the unconstrained PF                                                               & \cite{Aguirre2007}   & Set-Distribution \\
		                             & corr\_obj                  & correlation between objective values                                                                    & \cite{Liefooghe2019} & evolvability     \\
		                             & mean\_f                    & Average of unconstrained ranks                                                                          & \cite{Liefooghe2021} & y-distribution   \\
		                             & std\_f                     & Standard deviation of unconstrained ranks                                                               & \cite{Mersmann2011}  & y-distribution   \\
		                             & max\_f                     & Maximum of unconstrained ranks                                                                          & \cite{Mersmann2011}  & y-distribution   \\
		                             & skew\_f                    & Skewness of unconstrained ranks                                                                         & \cite{Mersmann2011}  & y-distribution   \\
		                             & kurt\_f                    & Kurtosis of unconstrained ranks                                                                         & \cite{Mersmann2011}  & y-distribution   \\
		                             & kurt\_avg                  & Average of objectives kurtosis                                                                          & \cite{Mersmann2011}  & y-distribution   \\
		                             & kurt\_min                  & Minimum of objectives kurtosis                                                                          & \cite{Mersmann2011}  & y-distribution   \\
		                             & kurt\_max                  & Maximum of objectives kurtosis                                                                          & \cite{Mersmann2011}  & y-distribution   \\
		                             & kurt\_rnge                 & Range of objectives kurtosis                                                                            & \cite{Mersmann2011}  & y-distribution   \\
		                             & skew\_avg                  & Average of objectives skewness                                                                          & \cite{Mersmann2011}  & y-distribution   \\
		                             & skew\_min                  & Minimum of objectives skewness                                                                          & \cite{Mersmann2011}  & y-distribution   \\
		                             & skew\_max                  & Maximum of objectives skewness                                                                          & \cite{Mersmann2011}  & y-distribution   \\
		                             & skew\_rnge                 & Range of objectives skewness                                                                            & \cite{Mersmann2011}  & y-distribution   \\
		                             & f\_mdl\_r2                 & Adjusted coefficient of determination of a linear regression model for varibles and unconstrained ranks & \cite{Mersmann2011}  & variable scaling \\
		                             & f\_range\_coeff            & Difference between maximum and minimum of the absolute value of the linear model coefficients           & \cite{Mersmann2011}  & variable scaling \\ \midrule
		\multirow{6}{*}{Random Walk} & dist\_f\_avg\_rws          & Average distance from neighbours in the objective space                                                 & \cite{Liefooghe2021} & evolvability     \\
		                             & dist\_f\_r1\_rws           & First autocorrelation coefficient of dist\_f\_avg\_rws                                                  & \cite{Liefooghe2021} & ruggedness       \\
		                             & dist\_f\_dist\_x\_avg\_rws & Ratio of dist\_f\_avg\_rws to dist\_x\_avg\_rws                                                         & \cite{Liefooghe2021} & evolvability     \\
		                             & dist\_f\_dist\_x\_avg\_r1  & First autocorrelation coefficient of dist\_f\_dist\_x\_avg\_rws                                         & \cite{Liefooghe2021} & ruggedness       \\
		                             & nuhv\_avg\_rws             & Average unconstrained hypervolume-value of neighborhood’s solutions                                     & \cite{Liefooghe2019} & evolvability     \\
		                             & nuhv\_r1\_rws              & First autocorrelation coefficient of nuhv\_avg\_rws                                                     & \cite{Liefooghe2019} & ruggedness       \\ \bottomrule
	\end{tabular}
\end{table*}

\begin{table*}[!t]
	\centering
	\caption{The features used to characterize the Violation Landscape of CMOP. The proposed features marked as New, while the (*) indicates that the feature has been modified to characterize CMOP.}
	\label{tab:features2}
	\scriptsize
	\begin{tabular}{llp{9cm}lp{2cm}}
		\toprule
		Type                          & \textbf{Feature}           & \textbf{Description}                                                                           & \textbf{Source}        & \textbf{Focus}   \\ \midrule
		\multirow{6}{*}{Global}       & min\_cv                    & Minimum of constraints violations                                                                                               & \cite{Mersmann2011} *  & y-distribution   \\
		                              & skew\_cv                   & Skewness of constraints violations                                                             & \cite{Mersmann2011} *  & y-distribution   \\
		                              & kurt\_cv                   & Kurtosis of constraints violations                                                             & \cite{Mersmann2011} *  & y-distribution   \\
		                              & cv\_mdl\_r2                & Adjusted coefficient of determination of a linear regression model for varibles and violations & \cite{Mersmann2011} *  & variable scaling \\
		                              & cv\_range\_coeff           & Difference between maximum and minimum of the absolute value of the linear model coefficients  & \cite{Mersmann2011} *  & variable scaling \\
		                              & dist\_c\_corr              & Violation-distance correlation                                                                 & \cite{Jones1995} *     & deception        \\ \midrule
		\multirow{10}{*}{Random Walk} & dist\_c\_avg\_rws          & Average distance from neighbours in the constraints space                                      & \cite{Liefooghe2021} * & evolvability     \\
		                              & dist\_c\_r1\_rws           & first autocorrelation coefficient of dist\_c\_avg\_rws                                         & \cite{Liefooghe2021} * & ruggedness       \\
		                              & dist\_c\_dist\_x\_avg\_rws & Ratio of dist\_c\_avg\_rws to dist\_x\_avg\_rws                                                & \cite{Liefooghe2021} * & evolvability     \\
		                              & dist\_c\_dist\_x\_r1\_rws  & First autocorrelation coefficient of dist\_c\_dist\_x\_avg\_rws                                & \cite{Liefooghe2021} * & ruggedness       \\
		                              & ncv\_avg\_rws              & Average single solution's violation-value                                                      & New                    & evolvability     \\
		                              & ncv\_r1\_rws               & first autocorrelation coefficient of ncv\_avg\_rws                                             & New                    & ruggedness       \\
		                              & nncv\_avg\_rws             & Average neighborhood’s violation-value                                                         & New                    & evolvability     \\
		                              & nncv\_r1\_rws              & first autocorrelation coefficient of nncv\_avg\_rws                                            & New                    & ruggedness       \\
		                              & bncv\_avg\_rws             & Average violation-value of neighborhood’s non-dominated solutions                              & New                    & evolvability     \\
		                              & bncv\_r1\_rws              & first autocorrelation coefficient of bncv\_avg\_rws                                            & New                    & ruggedness       \\ \bottomrule
	\end{tabular}
\end{table*}

\begin{table*}[!t]
	\centering
	\caption{The features used to characterize the Multi-Objectives-Violation Landscape of CMOP. The proposed features marked as New, while the (*) indicates that the feature has been modified to characterize CMOP.}
	\label{tab:features1}
	\scriptsize
	\begin{tabular}{llp{9cm}lp{2cm}}
		\toprule
		Type                          & \textbf{Feature}              & \textbf{Description}                                                        & \textbf{Source}         & \textbf{Focus}                     \\ \midrule
		\multirow{19}{*}{Global}      & fsr                           & Feasibility ratio                                                           & \cite{Malan2015}        & Set-Cardinality                    \\
		                              & po\_n                         & Proportion of PO solutions                                                  & \cite{Knowles2003}      & Set-Cardinality                    \\
		                              & hv                            & Hypervolume-value of the PF                                                 & \cite{Aguirre2007}      & Set-Distribution                   \\
		                              & cpo\_upo\_n                   & Proportion of PF to unconstrained PF                                        & New                     & PF and UPF correlation             \\
		                              & hv\_uhv\_n                    & Proportion of HV to unconstrained HV                                        & New                     & PF and UPF correlation             \\
		                              & GD\_cpo\_upo                  & distance between PF and unconstrained PF                                    & New                     & PF and UPF correlation             \\
		                              & cover\_cpo\_upo               & Proportion of unconstrained PF covered by PF                                & New                     & PF and UPF correlation             \\
		                              & corr\_cobj\_min               & Minimum constraints and objectives correlation                              & \cite{Vodopija2021}     & evolvability                       \\
		                              & corr\_cobj\_max               & Maximum constraints and objectives correlation                              & \cite{Vodopija2021}     & evolvability                       \\
		                              & corr\_cf                      & Constraints and ranks correlation                                           & \cite{Malan2015} *      & evolvability                       \\
		                              & piz\_ob\_min                  & Minimum proportion of solutions in ideal zone per objectives                & \cite{Malan2015} *      & Optima isolation                   \\
		                              & piz\_ob\_max                  & Maximum proportion of solutions in ideal zone per objectives                & \cite{Malan2015} *      & Optima isolation                   \\
		                              & piz\_f                        & Proportion of solutions in ideal zone                                       & \cite{Malan2015} *      & Optima isolation                   \\
		                              & ps\_dist\_max                 & Maximum distance across PS                                                  & \cite{Knowles2003}      & PS connectivity                    \\
		                              & ps\_dist\_mean                & Average distance across PS                                                  & \cite{Liefooghe2013b}   & PS connectivity                    \\
		                              & ps\_dist\_iqr\_mean           & Average difference between 75th and 25th percentiles of distances across PS & \cite{Liefooghe2013b}   & PS connectivity                    \\
		                              & pf\_dist\_max                 & Maximum distance across PF                                                  & \cite{Esty2022}         & PF discontinuouty                  \\
		                              & pf\_dist\_mean                & Average distance across PF                                                  & \cite{Esty2022}         & PF discontinuouty                  \\
		                              & pf\_dist\_iqr\_mean           & Average difference between 75th and 25th percentiles of distances across PF & \cite{Esty2022}         & PF discontinuouty                  \\  \midrule
		\multirow{21}{*}{Random Walk} & sup\_avg\_rws                 & Average proportion of neighbors dominating the current solution             & \cite{Liefooghe2019}    & evolvability                       \\ 
		                              & sup\_r1\_rws                  & First autocorrelation coefficient of sup\_avg\_rws                          & \cite{Liefooghe2019}    & ruggedness                         \\
		                              & inf\_avg\_rws                 & Average proportion of neighbors dominated by the current solution           & \cite{Liefooghe2019}    & evolvability                       \\
		                              & inf\_r1\_rws                  & First autocorrelation coefficient of inf\_avg\_rws                          & \cite{Liefooghe2019}    & ruggedness                         \\
		                              & inc\_avg\_rws                 & Average proportion of neighbors incomparable to the current solution        & \cite{Liefooghe2019}    & evolvability                       \\
		                              & inc\_r1\_rws                  & First autocorrelation coefficient of inc\_avg\_rws                          & \cite{Liefooghe2019}    & ruggedness                         \\
		                              & lnd\_avg\_rws                 & Average proportion of locally non-dominated solutions in the neighborhood   & \cite{Liefooghe2019}    & evolvability                       \\
		                              & lnd\_r1\_rws                  & First autocorrelation coefficient of lnd\_avg\_rws                          & \cite{Liefooghe2019}    & ruggedness                         \\
		                              & dist\_x\_avg\_rws             & Average distance from neighbours in the variable space                      & \cite{Liefooghe2021}    & evolvability                       \\
		                              & dist\_x\_r1\_rws              & First autocorrelation coefficient of dist\_x\_avg\_rws                      & \cite{Liefooghe2021}    & ruggedness                         \\
		                              & dist\_f\_c\_avg\_rws          & Average distance from neighbours in the objective-constraints space         & \cite{Liefooghe2021}  * & evolvability                       \\
		                              & dist\_f\_c\_r1\_rws           & First autocorrelation coefficient of dist\_f\_c\_avg\_rws                   & \cite{Liefooghe2021}  * & ruggedness                         \\
		                              & dist\_f\_c\_dist\_x\_avg\_rws & Ratio of dist\_f\_c\_avg\_rws to dist\_x\_avg\_rws                          & \cite{Liefooghe2021}  * & evolvability                       \\
		                              & dist\_f\_c\_dist\_x\_avg\_r1  & First autocorrelation coefficient of dist\_f\_c\_dist\_x\_avg\_rws          & \cite{Liefooghe2021}  * & ruggedness                         \\
		                              & nhv\_avg\_rws                 & Average hypervolume-value of feasible neighborhood’s solutions              & \cite{Liefooghe2019} *  & evolvability                       \\
		                              & nhv\_r1\_rws                  & First autocorrelation coefficient of nhv\_avg\_rws                          & \cite{Liefooghe2019} *  & ruggedness                         \\
		                              & bhv\_avg\_rws                 & Average hypervolume-value of neighborhood’s non-dominated solutions         & \cite{Liefooghe2019} *  & evolvability                       \\
		                              & bhv\_r1\_rws                  & First autocorrelation coefficient of bhv\_avg\_rws                          & \cite{Liefooghe2019} *  & ruggedness                         \\
		                              & nfronts\_avg\_rws             & Average number of ranks                                                     & New                     & evolvability                       \\
		                              & nfronts\_r1\_rws              & first autocorrelation coefficient of nfronts\_avg\_rws                      & New                     & ruggedness                         \\
		                              & rfbx\_rws\_avg                & Average ratio of feasible boundary crossings                                & \cite{Malan2015}        & Dispersion of the feasible regions \\ \bottomrule
	\end{tabular}
\end{table*}

\subsubsection{Multi-Objective Landscapes}\label{subsec:MOLandscape}

In multi-objective optimization, we are dealing with multiple fitness functions and set of optimal solutions. Therefore, Verel \etal~\cite{Verel2011} adjusted the elements of the fitness landscape to include multi-objective concepts by redefining $X$ as a set of solutions-sets, $f$ as multi-objective quality indicators, while $N$ depends on relations between solutions-sets. A set of features to characterize the multi-objective Landscape have been adopted from the literature and summarized in Table~\ref{tab:features3}.

\subsubsection{Violation Landscapes}\label{subsec:CVLandscape}

For characterizing constrained optimization problems, Malan \etal~\cite{Malan2015} introduced the violation landscape, which uses the same elements as a fitness landscape, but replaces $f$ by a violation function. The violation function quantifies how much a solution violates the problem constraints. Here, we use the norm of constraints violation vector as calculated in Equation~\ref{eq:CV}.

We propose six features to measure ruggedness and evolvability. From a random walk, we propose calculating average and first auto-correlation coefficient of the following: single solution's violation-value (ncv), its neighborhood total violation-value (nncv), violation-value of neighborhood’s non-dominated solutions (bncv).

In addition, a set of features from the fitness landscape domain has been modified to quantify the characteristics of the violation landscape. To measure the distribution of the violation function, we used minimum, kurtosis, and skewness features. To measure variable scaling, we used features from a linear regression model fitted to decision variables and their violation-value (cv{\_}mdl{\_}r2 and cv{\_}range{\_}coeff)~\cite{Mersmann2011}. To characterize the deception in the violation landscape, violation-distance correlation (dist{\_}c{\_}corr) has been calculated, which applied a similar concept to fitness-distance correlation~\cite{Jones1995}. We also modified the distance features from~\cite{Liefooghe2021} to calculate the average distance from neighbors in the constraints space (dist{\_}c), and its ratio to the average distance from neighbors in the variable space (dist{\_}c{\_}dis{\_}x). Average and first auto-correlation coefficient of the distance features have been collected. All the violation landscape features are presented in Table~\ref{tab:features2}.

\subsubsection{Multi-Objective-Violation Landscapes}\label{subsec:MOCVLndscape}

An important contribution of this paper is the introduction of multi-objective-violation landscape and features to characterize it. The multi-objective-violation landscape replaces Stadler's fitness function, $f$, by constraint domination principle~\cite{Deb2002} to measure the quality of solutions in the search space. Given two solutions $x$ and $y$, $x$ is said to have better quality or higher rank than $y$ if any of the following conditions is true:
\begin{inparaenum}[(a)]
	\item the solution $x$ is feasible and the solution $y$ is not;
	\item both solutions are infeasible but $x$ has smaller constraint violation norm; or
	\item $x$ and $y$ are both feasible or have similar constraint violation norm, but $x$ dominates $y$ w.r.t. objectives only.
\end{inparaenum}
Table~\ref{tab:features1} lists the features of the multi-objective-violation landscape.

Here, we propose six features. As described in~\ref{subsec:bench}, the relationship between unconstrained non-dominated solution set (upo) and constrained one (cpo) may cause some difficulty; therefore, four new features have been proposed to quantify this relationship. The first feature measures the relation size between the two sets by computing the proportion of cpo to upo (cpo{\_}upo{\_}n), while the second feature express the dominance relation between the two sets or in another word, the proportion of $UPF$ covered by $PF$ (cover{\_}cpo{\_} upo), the value of 1 means that all points in cpo are equal to points in upo, while 0 means none of the cpo is part of the upo~\cite{Zitzler1998b}. In addition, the proportion of hv to uhv (hv{\_}uhv{\_}n) is calculated to compare the volume of the objective space covered by the two sets. The last one is the distance between $PF$ and unconstrained $PF$ (GD{\_}cpo{\_}upo), which has been computed by using the generational distance metric~\cite{Lamont1999} in order to approximate the location of the $PF$ compared to the unconstrained one. Moreover, the average number of fronts (nfronts) in neighborhood and its first auto-correlation coefficient has been calculated as measures of evolvability and ruggedness of the landscape.

In order to include multi-objective and constraints concepts together, the following features from the multi-Objective or the violation landscapes are modified. In~\cite{Malan2015}, two features have been proposed to express the relation between the objectives and constraints functions in COPs, those are: the correlation between objective and constraints violation (corr{\_}c), and the proportion of solution in ideal zone (piz). Vodopija \etal~\cite{Vodopija2021} proposed using the correlations between each objective and the constraint violation, focusing on the  minimum and maximum correlation coefficients. Here, our new feature measures the correlation between constraints violation norm and solution rank based on the constraint domination principle (corr{\_}cf) as well. We also apply (piz) for the first time to calculate the minimum and maximum proportion of solutions in ideal zone per objective (piz{\_}ob), as well as calculating the proportion of ranked solutions in ideal zone (piz{\_}f). Furthermore, from a sample collected by random walk, we extend the use of distance features from multi-objective landscape~\cite{Liefooghe2021} to calculate the average distance from neighbors in the multi-objective-violation space (dist{\_}f{\_}c), and its ratio to the average distance from neighbors in the variable space (dist{\_}f{\_}c{\_}dis{\_}x). For features based on $HV$ of neighborhoods, we propose calculating the $HV$ covered by the feasible set (nhv) and the $HV$ covered by the non-dominated set (bhv).

\subsection{Algorithm Space}\label{subsec:CMOEA}

Specialized versions of multi-objective evolutionary algorithms (MOEAs) have been designed with constraints handling techniques, so called constrained multi-objective evolutionary algorithms (CMOEAs), to maintain the necessary balance between optimizing objectives and satisfying constraints in CMOP. There are three categories of CMOEAs~\cite{Tian2021}:

\begin{description}
	\item[Prioritize constraints:] methods of this category pressure the search toward feasible regions, however, algorithms may get trapped in a small part of the feasible region because of the bias toward infeasible solutions. Representative methods of this category include using the principle of constraint dominance such as NSGAII, DCMOEAD~\cite{Deb2002}, and ANSGAIII~\cite{Jian2014}, a relaxed version of constraint dominance such as $\epsilon$-constraint~\cite{Fan2019}. ECNSGAII and ECMOEAD apply an improved version of $\epsilon$-constraint which has been proposed in~\cite{Fan2019PPS}.
	\item[Consider objectives and constraints equally:] methods belong to this category treat constraints as part of the objective functions~\cite{Angantyr2003} by including static or dynamic penalty factor in the objectives~\cite{Jan2013} such as $\epsilon$-constraint dynamic penalty~\cite{Zhu2020MOEADDAE} which has been used in PECNSGAII and PECMOEAD. Other methods objectivize the constraints~\cite{Ning2017}, or switch between dominance relation to compare constraints and dominance to compare objectives by using stochastic ranking~\cite{Runarsson2020}, which has been implemented in SRNSGAII and SRCMOEAD, or use the status of the search such as CMOEA{\_}MS~\cite{Tian2021}. This approach provides good balance between exploring feasible and non-feasible regions, however, they may suffer in convergence. 
	\item[Hyper-strategies:] methods of this category use different strategies in different populations or stages. They aim to balance objectives and constraints by favoring one or both in each stage of the search or in different populations. For example, CTAEA~\cite{Chen2018} uses two archives, one to maintain convergence by optimizing both constraints and objectives, while the second archive is used to maintain diversity, and it considers optimizing objectives only. On the other hand, CCMO~\cite{Tian2021CCMO} uses two populations, one to solve the original CMOP and another to solve a helper problem derived from the original one. Another approach is to use multiple stages of the search, MOEADDAE~\cite{Zhu2020MOEADDAE} uses the first stage to push the search toward feasible solutions by prioritizing constraints and the second stage to favor objectives in order to escape local optima, whilst PPS~\cite{Fan2019PPS} pushes the search towards $UPF$, then, pulls it to the feasible region. ToP~\cite{Liu2019ToP} converts CMOP into COP in the first stage, then uses a CMOEA in the second stage.
\end{description}

\subsection{Performance Space}\label{subsec:performance}

The most commonly used performance indicator when optimizing CMOPs is Hypervolume ($HV$)~\cite{Zitzler2007,Abhishek2021},  which quantifies the volume of the objective space covered by $\widetilde{PF}$ and a reference point to measure $\widetilde{PF}$ convergence and distribution. The reference point, $r$, is a vector that has objective values worse than any values in the $\widetilde{PF}$. To overcome $HV$ bias, a common reference point $r=(1.1,\ldots,1.1)^{(T)}$ is used with the normalized $PF$ and objectives~\cite{Audet2018}. The larger the value of the $HV$, the better the approximation of the true $PF$.

We use a binary concept to define the `goodness' of the measured performance with respect to others~\cite{Yap2020}. We consider the performance of an algorithm as a `good' performance if the normalized HV is greater than zero and within 1\% of the best algorithm on the same instance.

\section{Experimental Setup}\label{sec:Exp}

We have used a total of 383 bi-objective instances to explore the characteristics of CMOPs and to study their impact on the performance of CMOEAs. Instances belong to the six benchmark suites described in Section~\ref{subsec:bench}, with $n=\in\left\{2, 5, 10\right\}$, except for CF, where $n=2$ cannot be used. For the DAS-CMOP suite, 15 instances were generated from each problem at each value of $n$ by varying the constraints parameters to adjust difficulty.

To extract the features, for each dimension, 30 samples sets were generated of size $n\times 10^{3}$. The average of features were then calculated. Global features used random sampling, while local features depend on random walks. A random walk of neighborhood size $N=(2\times n)+1$, length $(n/N)\times 10^{3}$, step size of 2\% of the range of the instance domain. Then, the features were processed using the Yeo-Johnson power transform method, which resulted in a distribution closer to Gaussian.

We have tested 15 algorithms, five from each category of the CHTs described in ~\ref{subsec:CMOEA}. NSGAII, ANSGAIII, CMOEA{\_}MS, CTAEA, CCMO, MOEADDAE, PPS, and ToP are available through the PlatEMO~\cite{PlatEMO} library, while DCMOEAD, ECNSGAII, ECMOEAD, PECNSGAII, PECMOEAD, SRNSGAII, and SRCMOEAD have been implemented as described in Section~\ref{subsec:CMOEA}. All MOEAD-based algorithms used Tchebycheff approach. We have used algorithms' default parameters. The population size set to be 200 with all instances, while number of evaluations is set to be $2\times10^{4}$ for $n=2$ instances and $5x10^{4}$ for $n=\left\{5,10\right\}$. For each algorithm and each instance, 30 independent runs were conducted. The average Max-Min normalized HV has been calculated for the performance metric, then, converted to binary performance as described in Section~\ref{subsec:performance}.

After collecting the meta-data, we filtered out features that were weakly correlated with algorithm performance. To do so, we removed those features that have an absolute value of the Pearson correlation less than 0.3 with all algorithms performances. Then, when the absolute value of Pearson correlation of two features was greater than 0.85, one of them was eliminated to reduce redundancy. We then used random forest regressor (RF) to keep only the features that were predictive of algorithms' performance. Hyperparameter tuning and 3-fold cross validation were used to build more accurate and stable RF models. Finally, to construct the instance space, we make use of the publicly available web tools in MATILDA~\cite{MATILDA}. 

\section{Results}\label{sec:results}

\subsection{Instance Space}\label{subsec:IS}

\begin{figure}[!t]
	\includegraphics[width=0.48\textwidth]{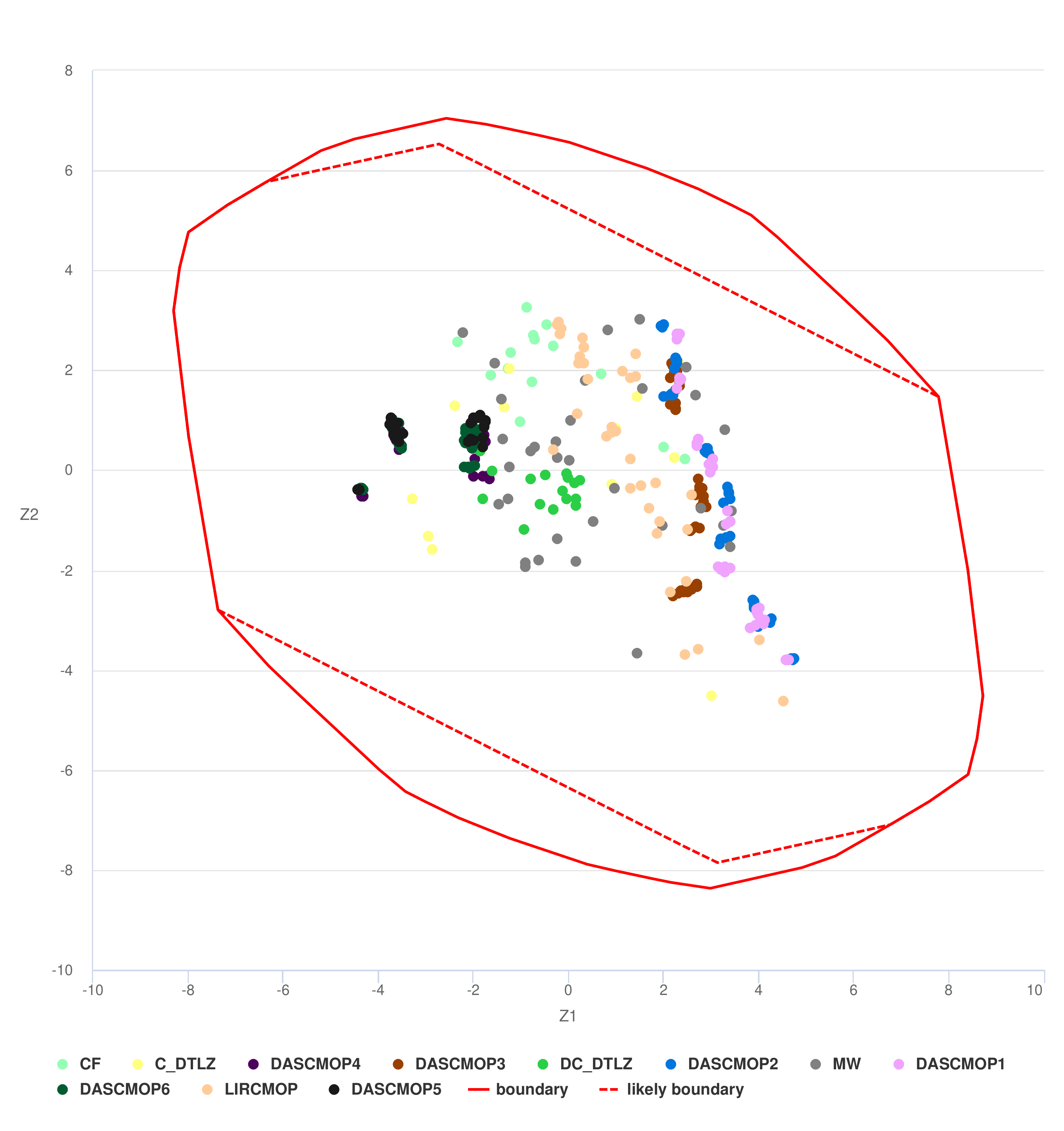}
	\caption{Distribution of the instances in 2D space by using the projection matrix in Equation~\eqref{eq:IS}. Instances are color-coded based on the source.}
	\label{fig:source}
\end{figure}

Figure~\ref{fig:source} illustrates the 2-dimensional CMOP instance space, where each instance is represented as a point. The location of each instance is defined by the following projection matrix: 
\begin{equation}
	\footnotesize
	\begin{bmatrix}[c]
		{z_{1}}\\
		{z_{2}}
	\end{bmatrix}=
	\begin{bmatrix}[r]
		-0.0682 & -0.2608 \\
		-0.0465 & 0.2616 \\
		0.1413 & -0.0689 \\
		-0.1132 & 0.0217 \\
		-0.2930 & -0.1596 \\
		0.2010 & 0.0430 \\
		0.2178 & -0.0309 \\
		0.2008 & -0.1819 \\
		-0.1996 & 0.0440 \\
		-0.3420 & 0.3035 \\
		0.3196 & -0.1020 \\
		0.2640 & 0.0873 \\
		0.1285 & 0.0726 \\
		-0.2986 & 0.1138 \\
		-0.2306 & 0.2422 \\
		0.1911 & -0.0884 \\
		0.1912 & 0.0413 \\
		-0.2418 & 0.0489 \\
		-0.0513 & 0.4075 \\
		-0.0465 & 0.3733 \\
		0.0397 & 0.2717 \\
		0.2087 & -0.1661 \\
		0.1462 & 0.0967
	\end{bmatrix}^{\top}
	\begin{bmatrix}[l]
		min{\_}cv \\
		skew{\_}cv \\
		pop{\_}cv{\_}mdl{\_}r2 \\
		pop{\_}cv{\_}range{\_}coeff \\
		dist{\_}c{\_}dist{\_}x{\_}avg{\_}rws \\
		nncv{\_}r1{\_}rws \\
		bncv{\_}r1{\_}rws \\
		upo{\_}n \\
		corr{\_}obj \\
		mean{\_}f \\
		skew{\_}f \\
		f{\_}mdl{\_}r2 \\
		f{\_}range{\_}coeff \\
		dist{\_}f{\_}dist{\_}x{\_}avg{\_}rws \\
		cpo{\_}upo{\_}n \\
		GD{\_}cpo{\_}upo \\
		hv \\
		corr{\_}cf \\
		piz{\_}ob{\_}min \\
		ps{\_}dist{\_}mean \\
		nhv{\_}avg{\_}rws \\
		bhv{\_}avg{\_}rws \\
		nhv{\_}r1{\_}rws
	\end{bmatrix}
	\label{eq:IS}
\end{equation}
\noindent which uses the features with the highest correlation with algorithm performance. An inspection of the figure reveals that most test suites are distributed over specific parts of the instance space. This is expected, as instances in the same suite usually share similar objective or constraint functions. For example, instances from DAS-CMOP1, DAS-CMOP2, and DAS-CMOP3 have similar constraint functions, while DC-DTLZ instances have either DTLZ1 or DTLZ3 objective functions. We observe that there is a paucity of instances in the bottom-left side of the space, which suggests a lack of diversity in some features. By observing the features' distribution in Figure~\ref{fig:features}, we note that there is a lack of instances with highly negative $corr{\_}cf$ and instances that have high $cpo{\_}upo{\_}n$. Also, there is high density of instances in some regions, attributed to the DAS-CMOP suite. We note that changing the constraints parameters of such problems does not have impact on the difficulty or diversity of instances. However, changing $n$ does have an impact.

\begin{figure}[!t]
	\includegraphics[width=0.48\textwidth]{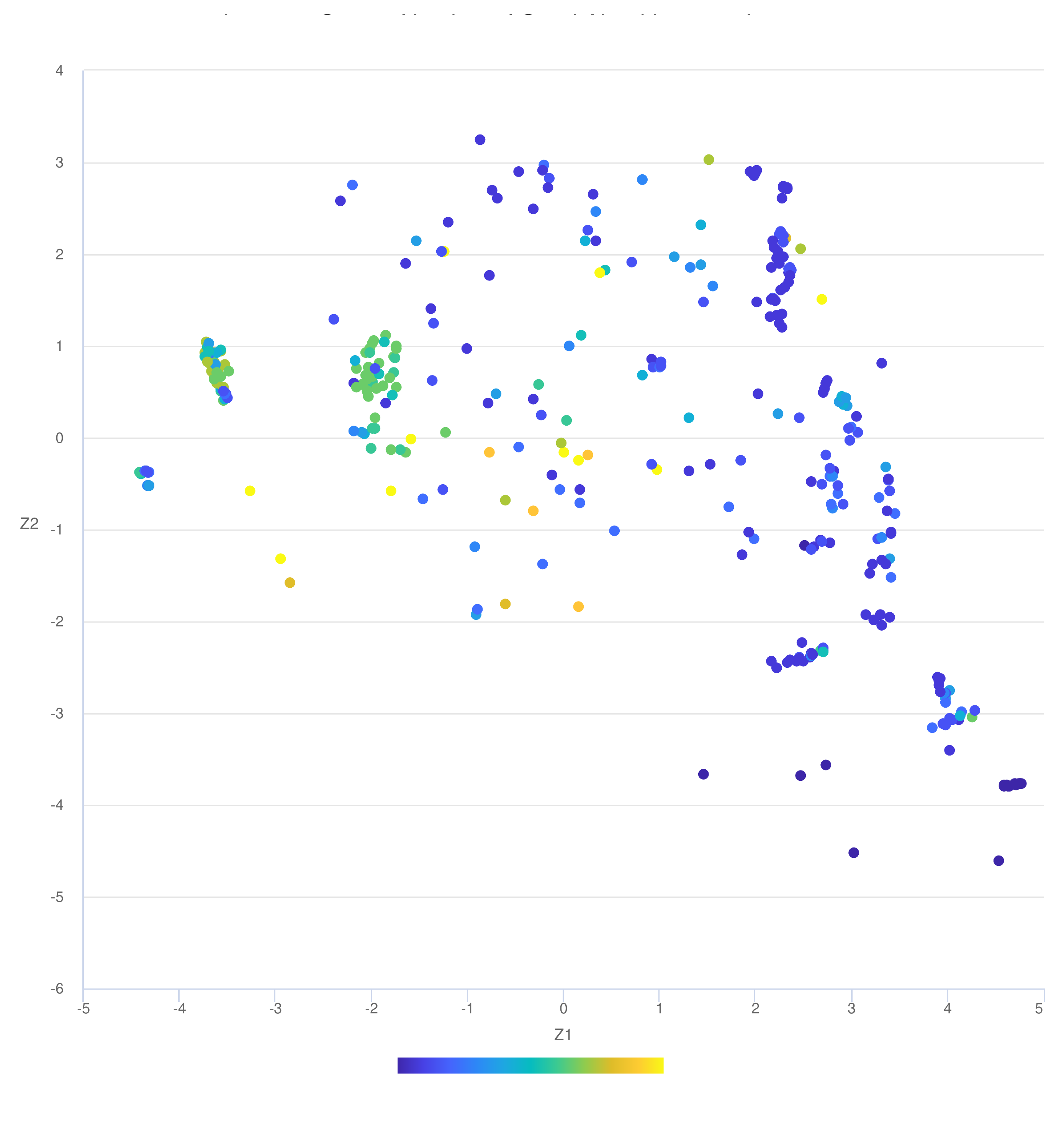}
	\caption{Number of algorithms performed well for each instance, where good performance means a normalized $HV$ within 1\% of the best algorithm. The color scale corresponds to the total number of algorithms. A color closer to dark blue means fewer number of algorithms performed well.}
	\label{fig:goodAlgo}
\end{figure}

Figure~\ref{fig:goodAlgo} illustrates the number of algorithms that performed well on each instance, as a proxy of their difficulty. Darker colors of the points in the plot corresponds to fewer algorithms performing well. The figure suggests that the instances in the left area are generally easier to solve by most of the algorithms, especially the instances in  the bottom-left of the plot.

\subsection{Algorithms Footprints}\label{subsec:footprints}

\begin{figure*}[!t]
	\subfloat[][NSGAII]{
		\includegraphics[width=0.24\textwidth]{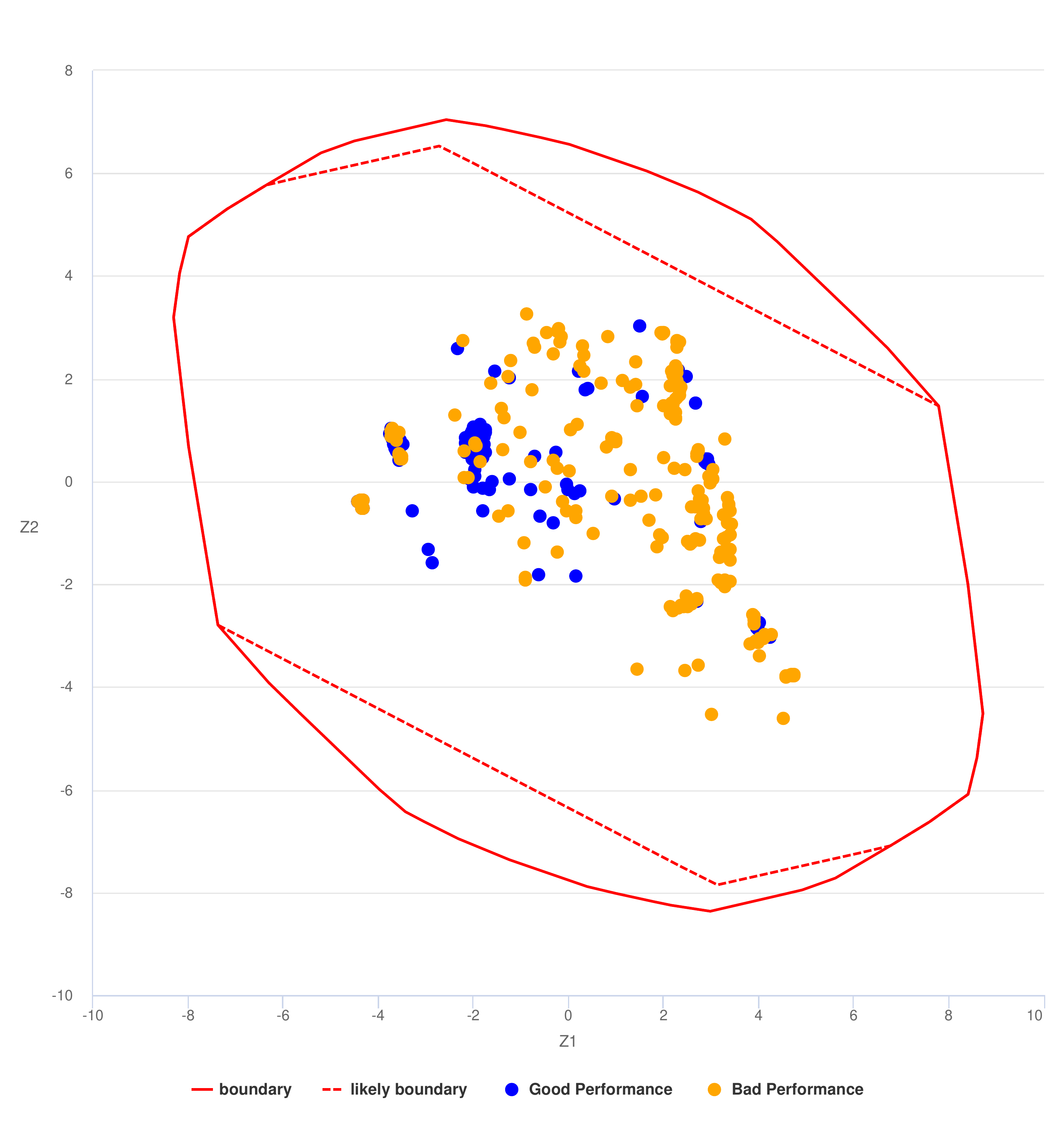}
		\label{fig:NSGAII}
	}%
	\subfloat[][PECMOEAD]{
		\includegraphics[width=0.24\textwidth]{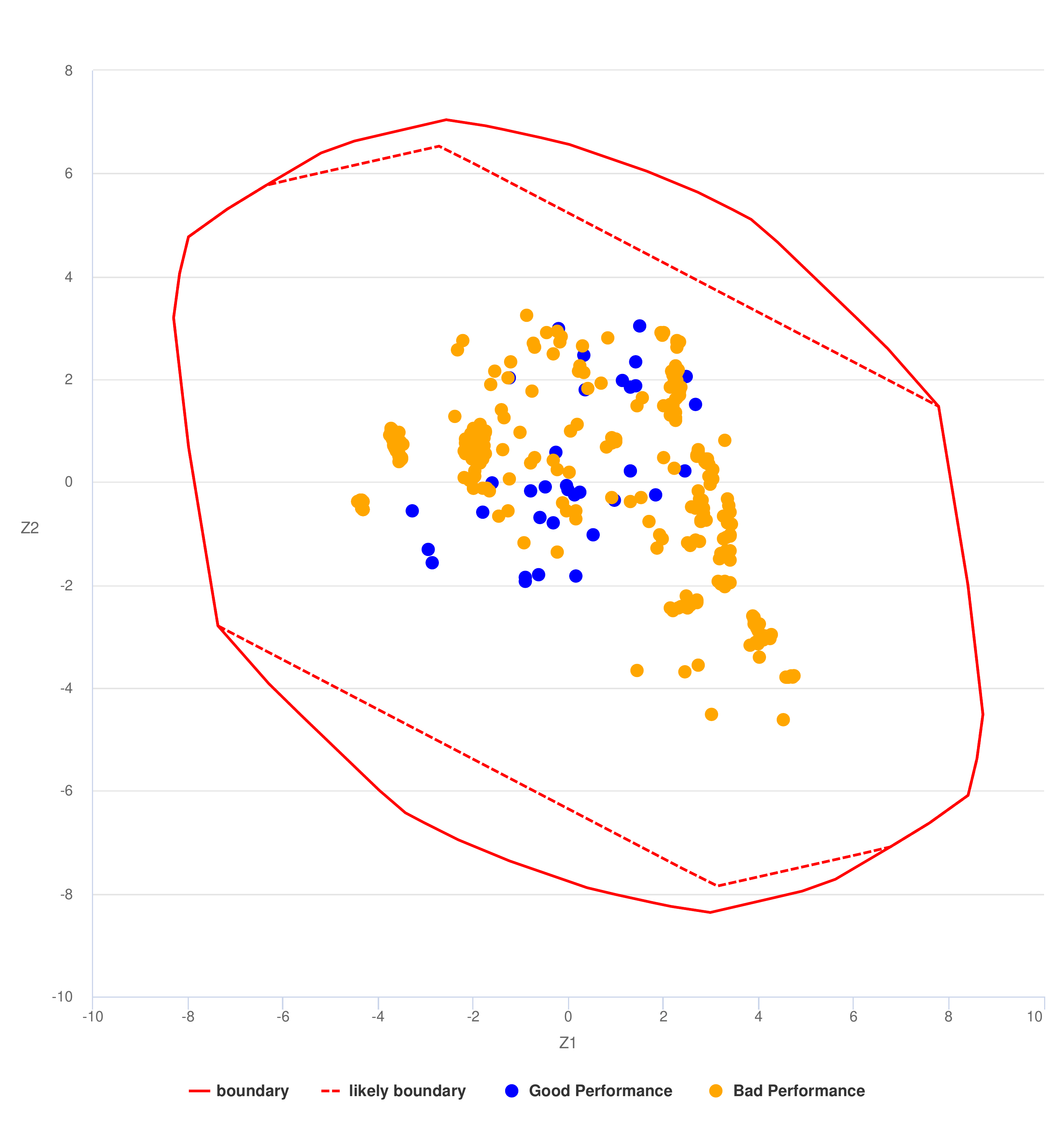}
		\label{fig:PECMOEAD}
	}%
	\subfloat[][MOEADDAE]{
		\includegraphics[width=0.24\textwidth]{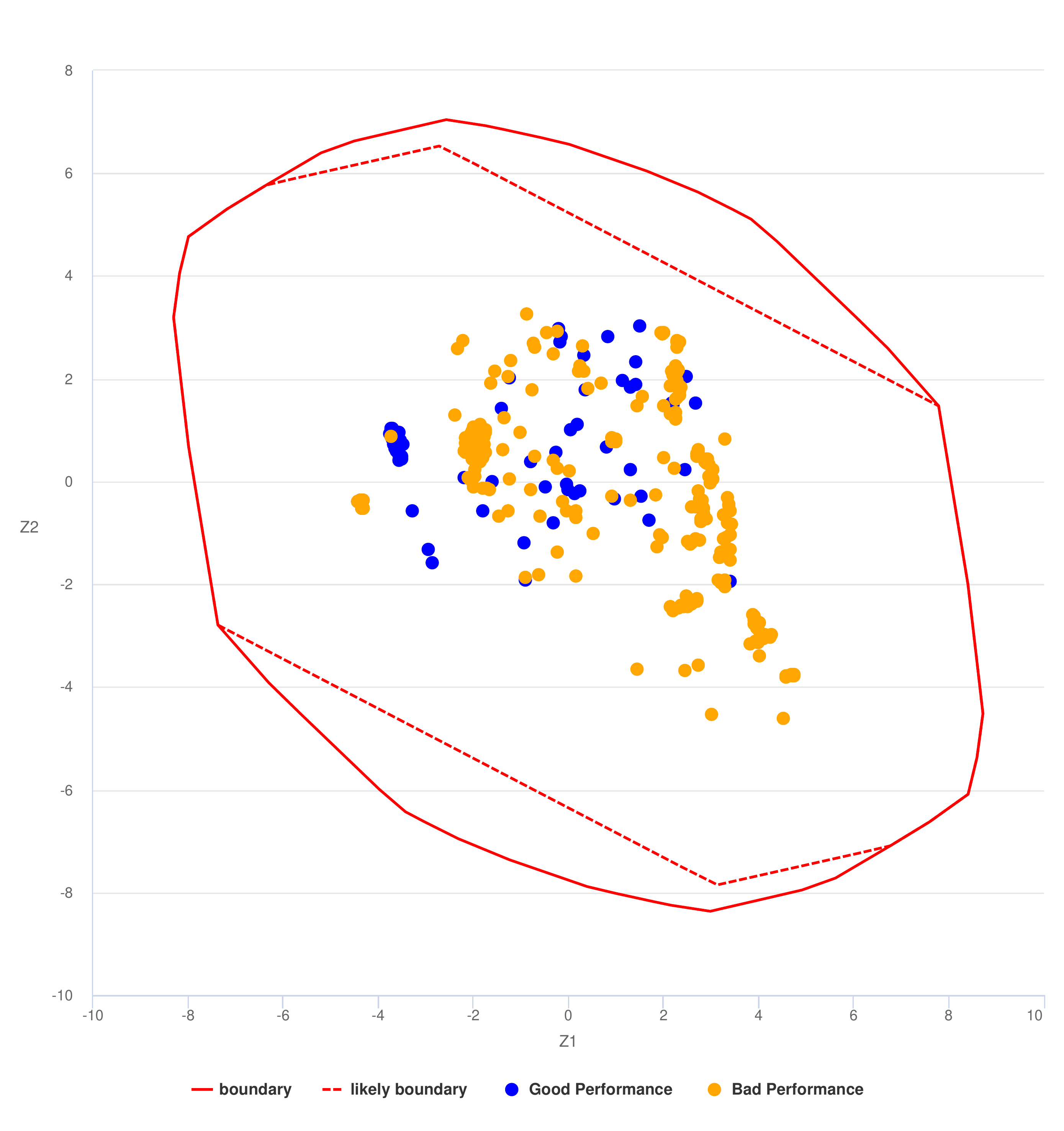}
		\label{fig:MOEADDAE}
	}%
	\subfloat[][CTAEA]{
		\includegraphics[width=0.24\textwidth]{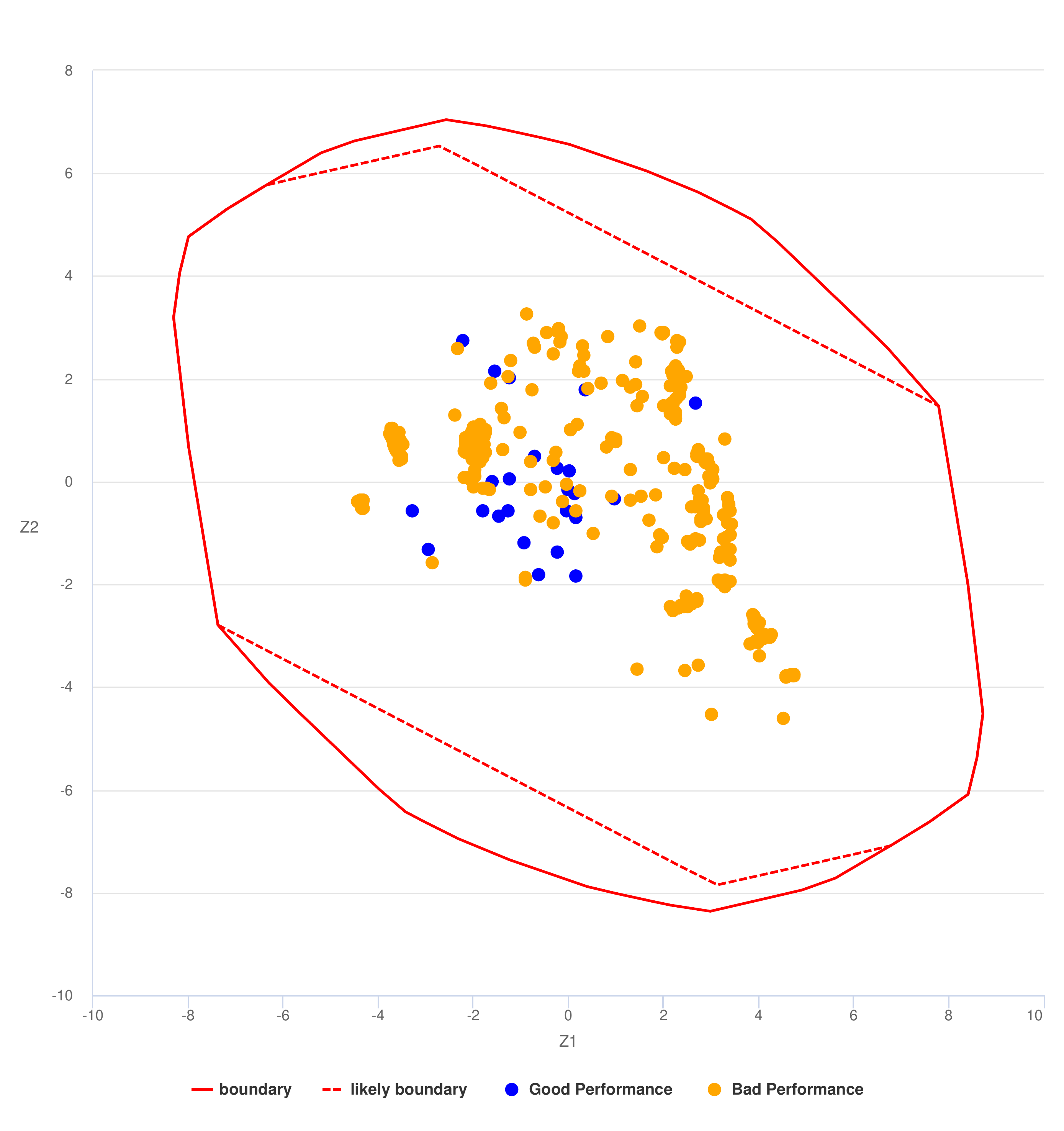}
		\label{fig:CTAEA}
	}\\%
	\subfloat[][ToP]{
		\includegraphics[width=0.24\textwidth]{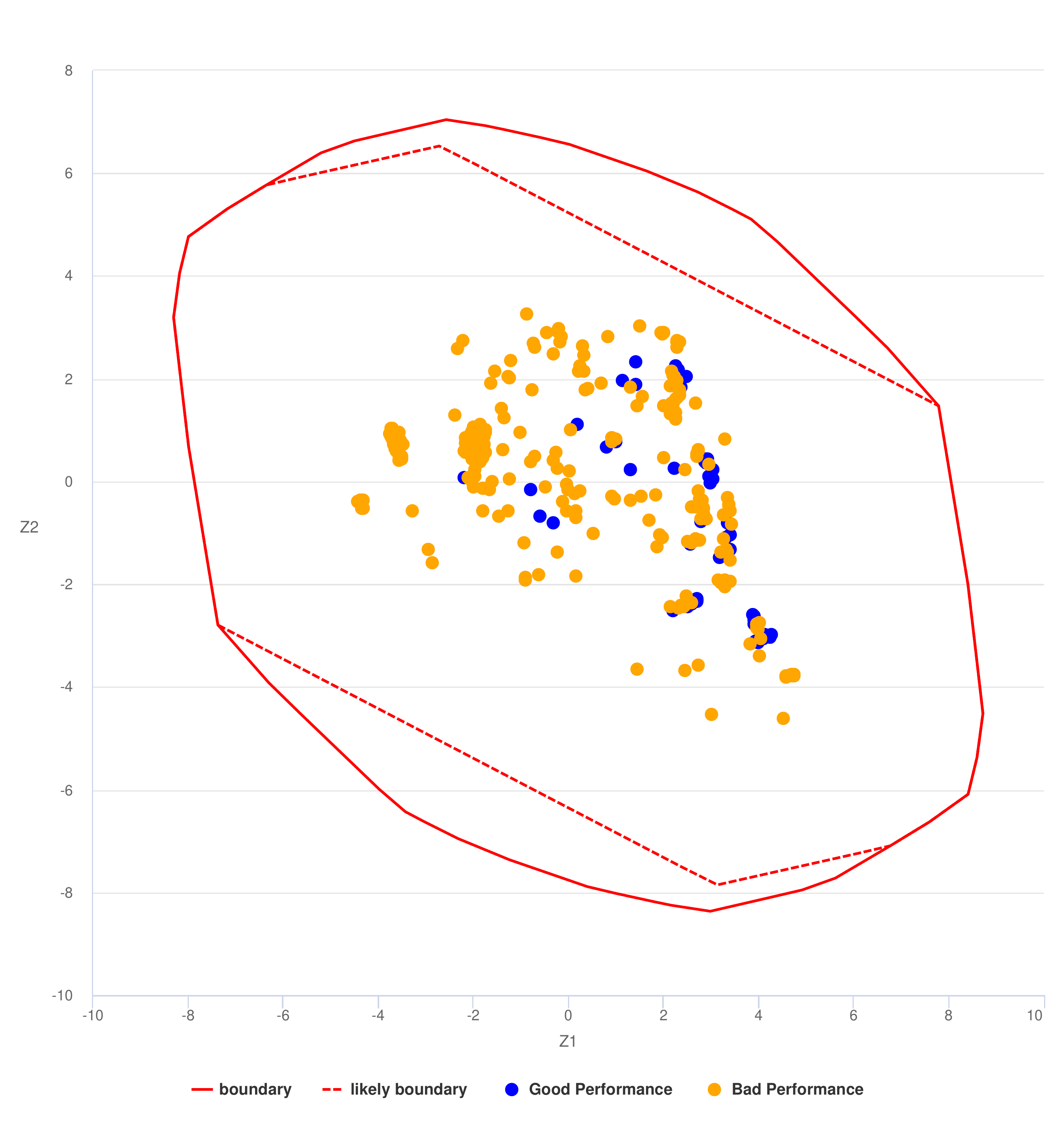}
		\label{fig:ToP}
	}%
	\subfloat[][CCMO]{
		\includegraphics[width=0.24\textwidth]{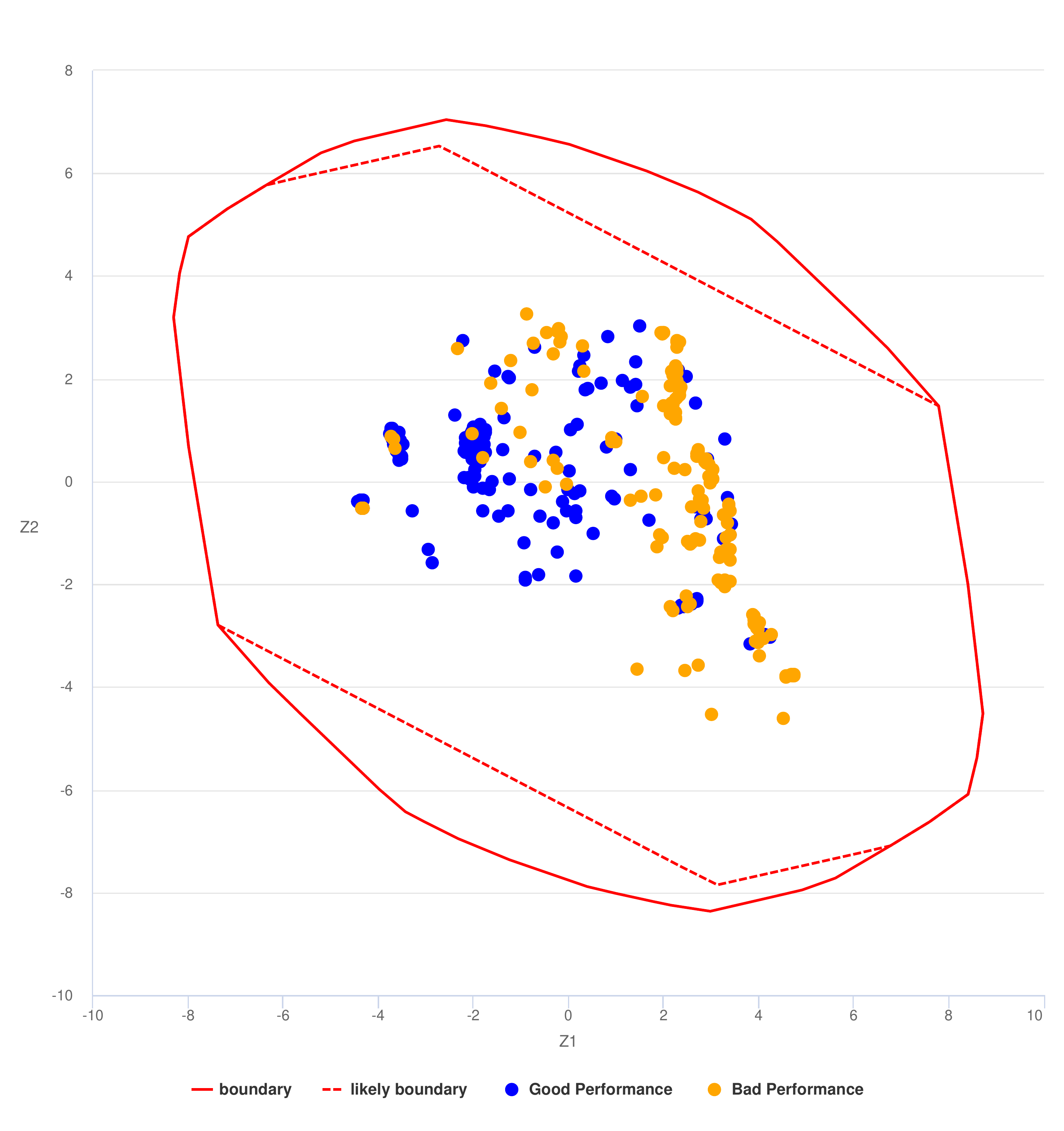}
		\label{fig:CCMO}
	}%
	\subfloat[][PPS]{
		\includegraphics[width=0.24\textwidth]{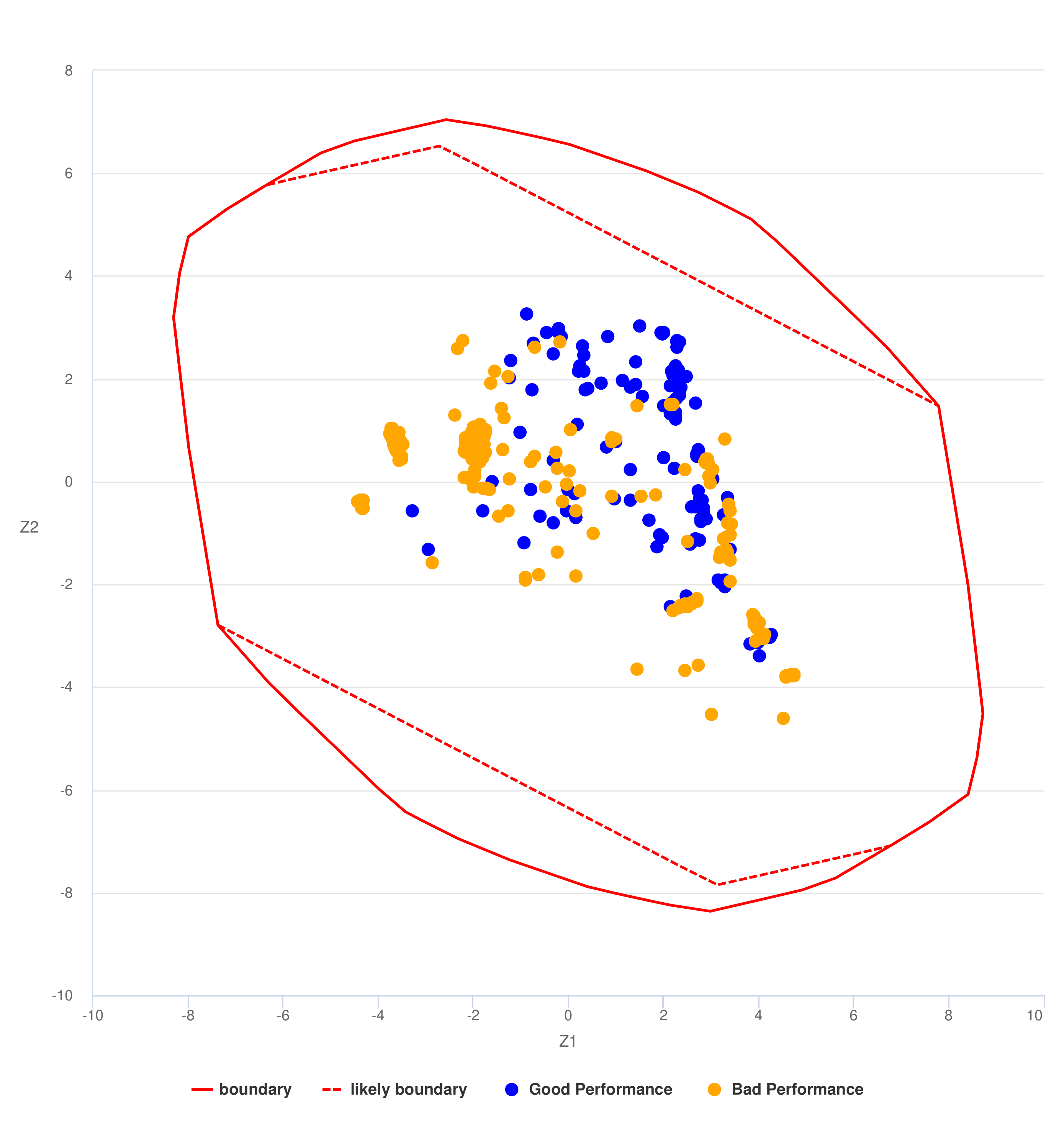}
		\label{fig:PPS}
	}
	\caption{Seven algorithms footprints in the projected instance space, where good performance means a normalized $HV$ within 1\% of the best algorithm.}
	\label{fig:afootprints}
\end{figure*}

Figure~\ref{fig:afootprints} shows the footprints of the algorithms in the instance space. An orange point means the algorithm performed badly compared to others on such instances, while blue represent good performance. We limit our analysis to only seven algorithms, as footprints reveal high similarity between many of them. ANSGAIII, NSGAII and MOEAD with the principle of constraint dominance, $\epsilon$-constraint, and stochastic ranking share similar footprints, as well as CMOEA{\_}MS which depends on the principle of constraint dominance but sometimes includes the constraint violation as an objective. The footprints of this group are represented by the footprints of NSGAII in~\ref{fig:NSGAII}, which shows that they are capable of providing relatively good performance in only a third of the instance space. Penalty based algorithms (PECNSGAII, PECMOEAD)~\ref{fig:PECMOEAD} have similar footprints; both of them matched the best algorithm in instances located near the origin or on the bottom-left area. Whilst, MOEADDAE~\ref{fig:MOEADDAE}, which uses two stages to relax the penalty factor, performed well in almost all the area covered by penalty based algorithms, and matched part of the first group footprints.

On the other hand, CTAEA, ToP, CCMO, and PPS have distinctive footprints. CTAEA and ToP have the lowest proportion of good performance, but they have different footprints. CTAEA~\ref{fig:CTAEA} seems only capable of providing high quality solutions in easy to solve instances, while ToP~\ref{fig:ToP} targeted instances that are rarely solved by previous algorithms. CCMO~\ref{fig:CCMO} and PPS~\ref{fig:PPS} appear to be the only two algorithms performed well in a wide area of the instance space. Moreover, they have almost opposite footprints. Both algorithms use two strategies to handle constraints, the first one is considering objectives only, but for the second CCMO uses the principle of constraint dominance while PPS uses $\epsilon$-constraint. CCMO uses the two strategies in parallel by having two populations, while PPS uses them sequentially by applying the first strategy for several generations, then, applying the second strategy.

\subsection{Features Impact}\label{subsec:importance}

\begin{figure*}[!t]
	\subfloat[][corr{\_}cf]{
		\includegraphics[width=0.32\textwidth]{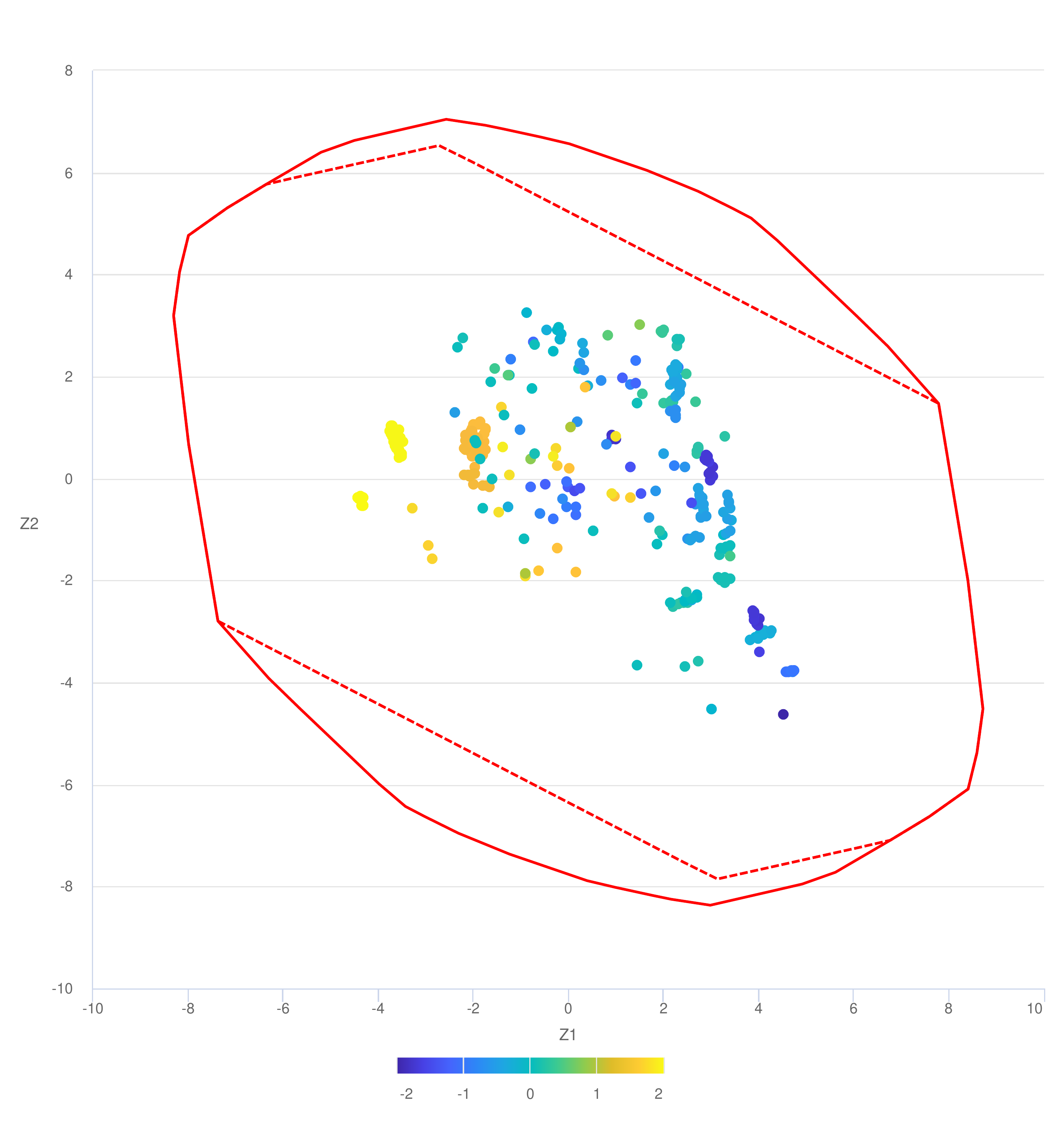}
		\label{fig:corr_cf}
	}%
	\subfloat[][cpo{\_}upo{\_}n]{
		\includegraphics[width=0.32\textwidth]{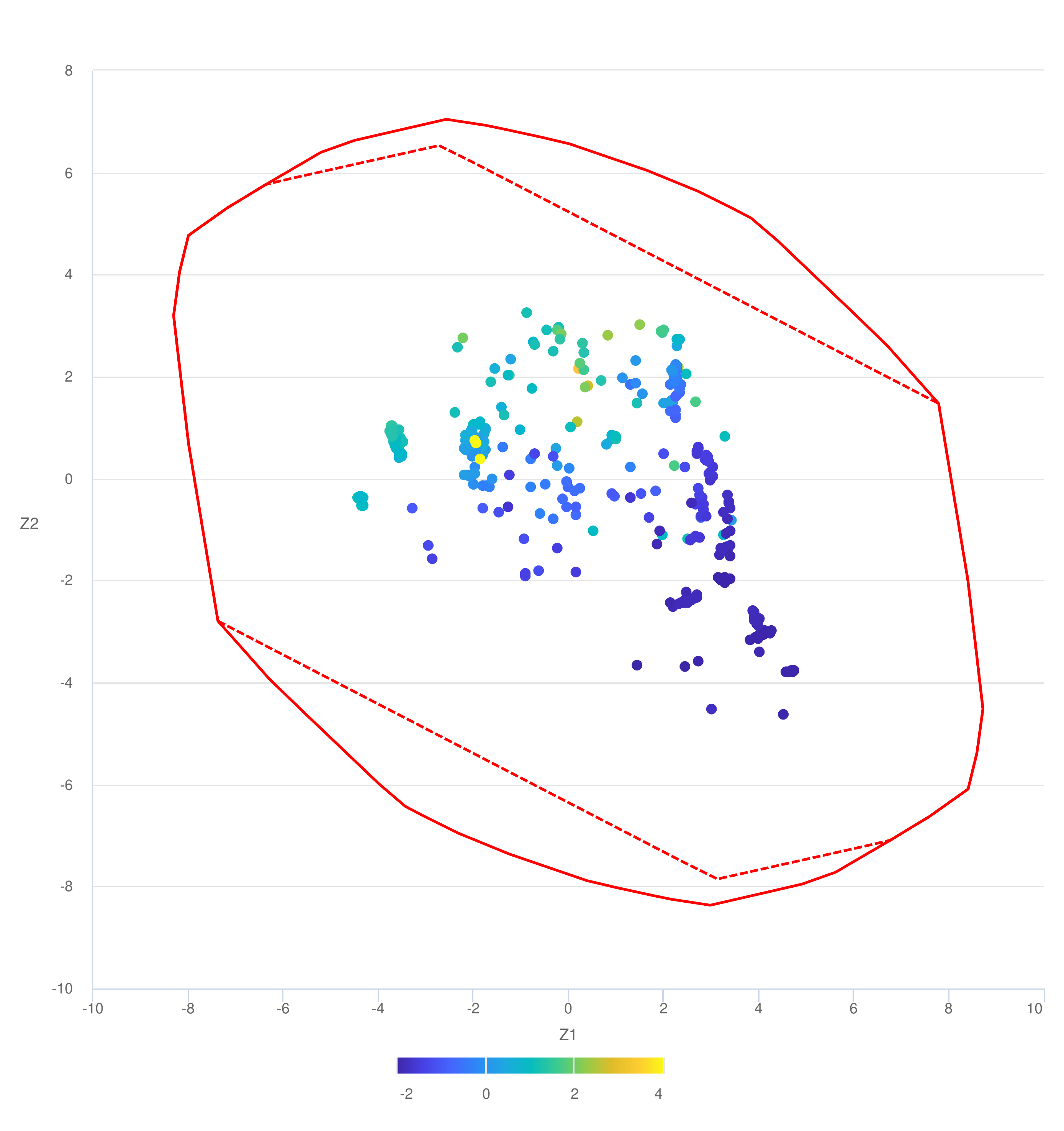}
		\label{fig:cpo_upo}
	}%
	\subfloat[][dist{\_}c{\_}dist{\_}x{\_}avg{\_}rws]{
		\includegraphics[width=0.32\textwidth]{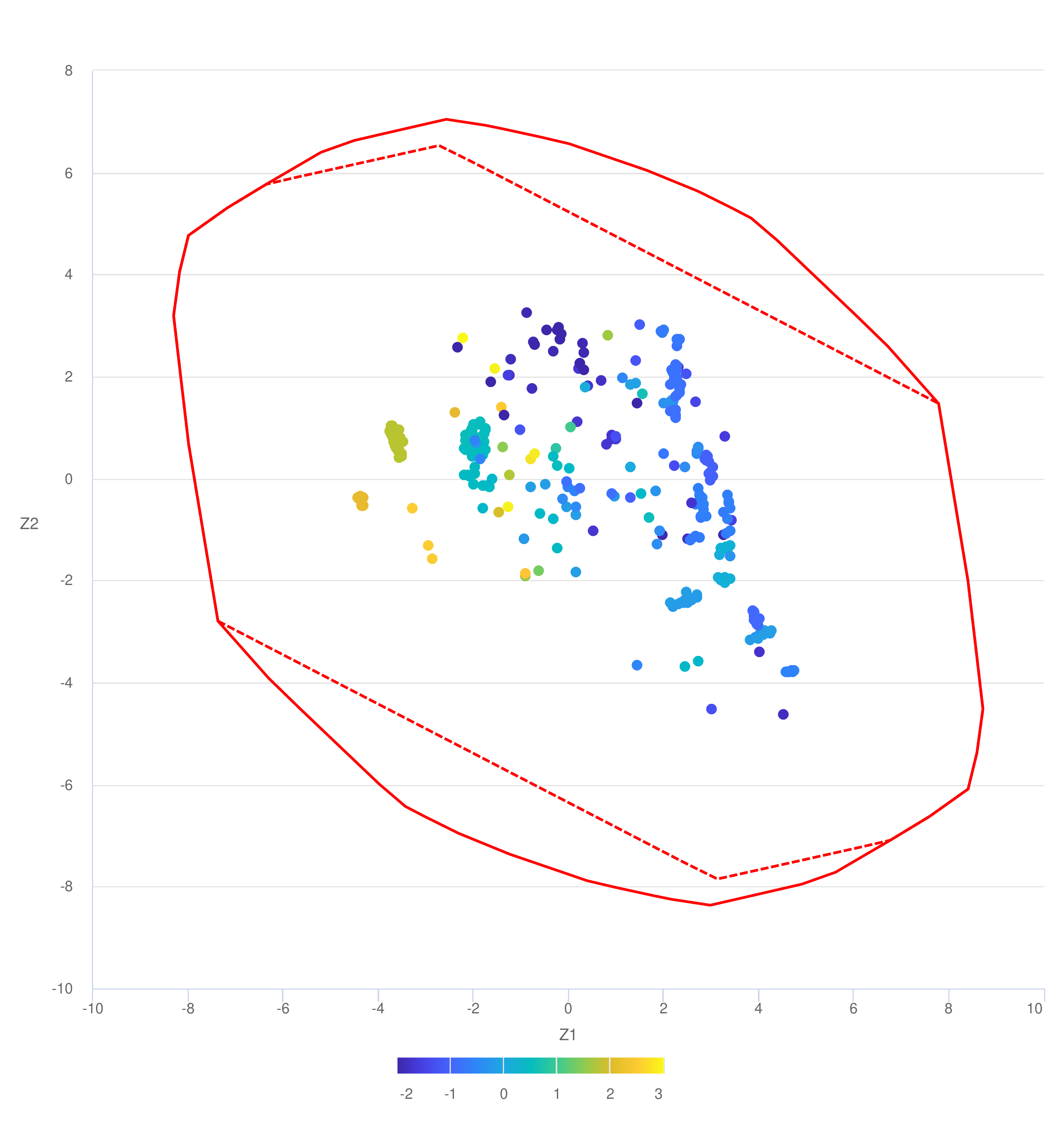}
		\label{fig:dist_c}
	}\\%
	\subfloat[][dist{\_}f{\_}dist{\_}x{\_}avg{\_}rws]{
		\includegraphics[width=0.32\textwidth]{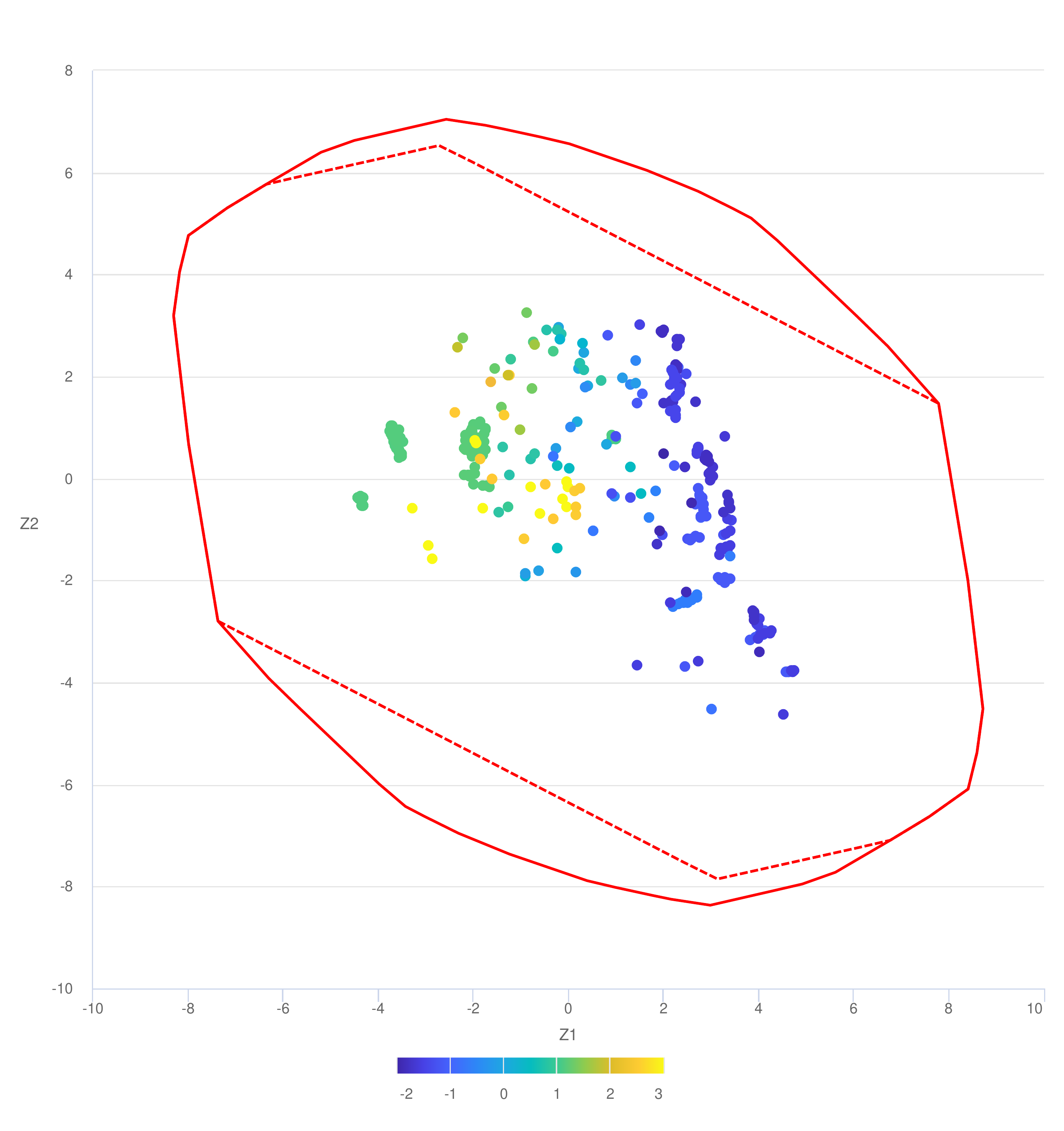}
		\label{fig:dist_f}
	}%
	\subfloat[][piz{\_}ob{\_}min]{
		\includegraphics[width=0.32\textwidth]{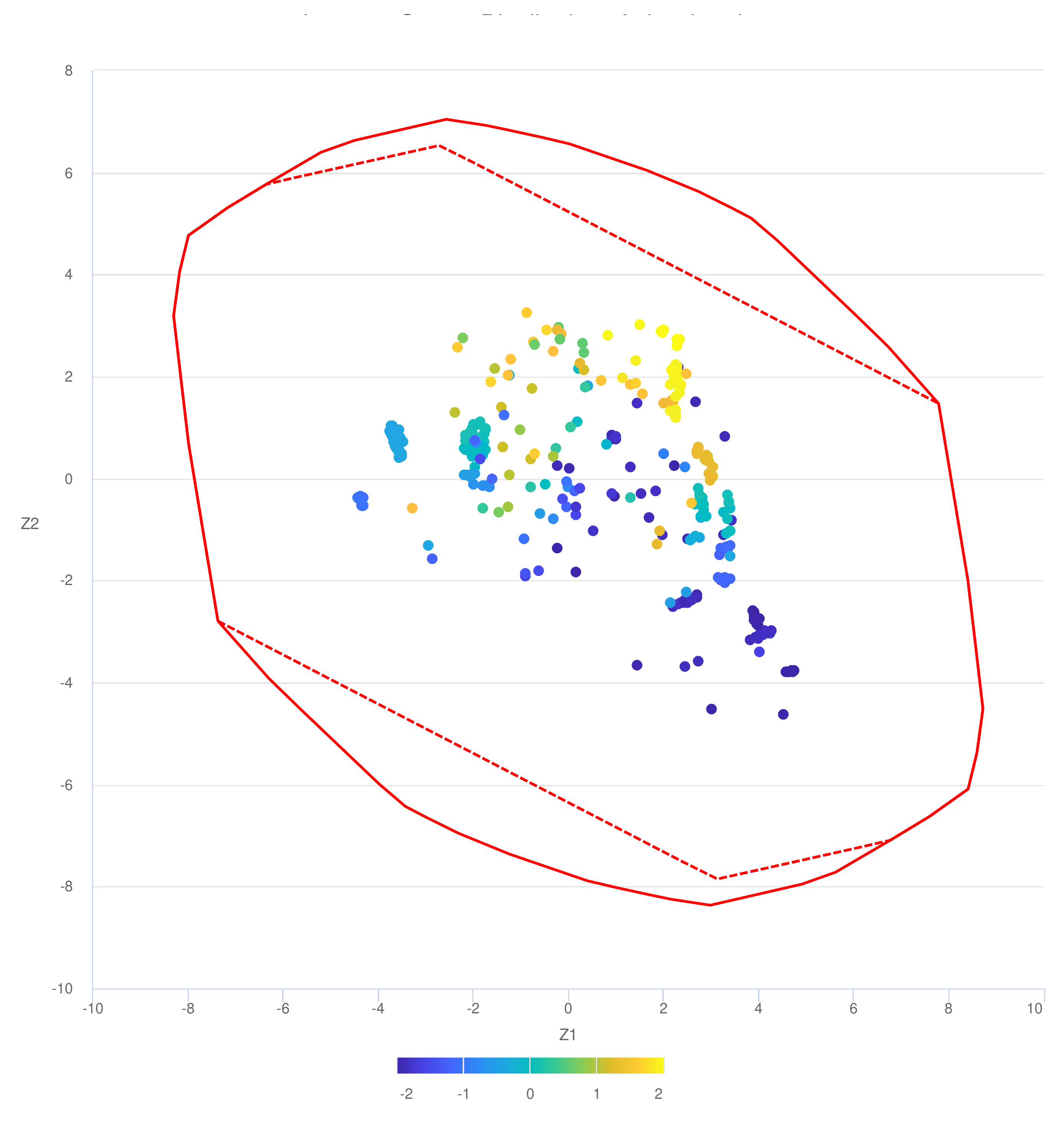}
		\label{fig:piz_ob}
	}%
	\subfloat[][nhv{\_}avg{\_}rws]{
		\includegraphics[width=0.32\textwidth]{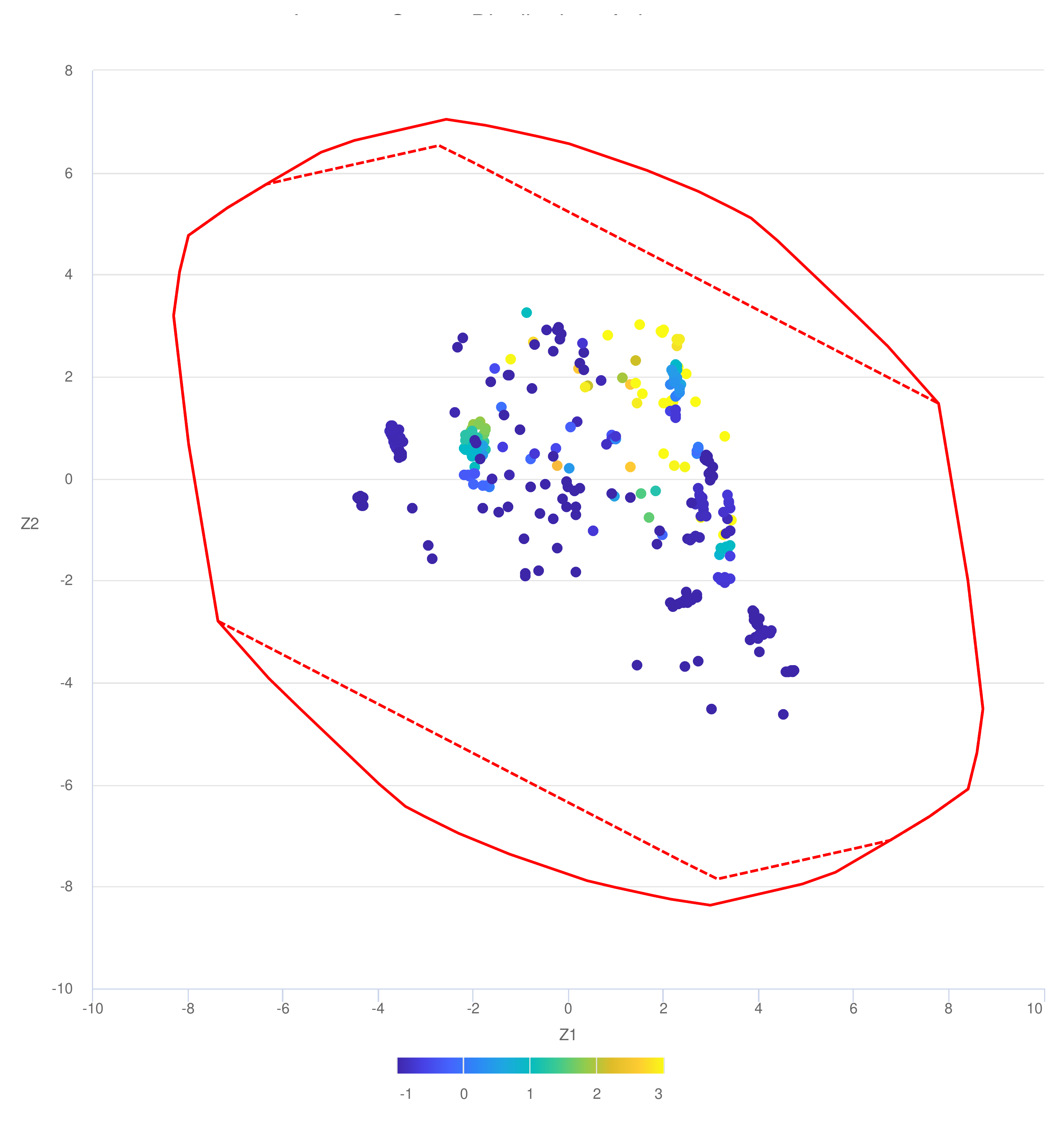}
		\label{fig:nhv_avg}
	}%
	\caption{Distribution of normalized subset of features in the projected instance space. The color scale corresponds to normalized feature values.}
	\label{fig:features}
\end{figure*}

Within the instance space, we can gain insights into an algorithm's strength and weakness by examining the distribution of features across the space. Here, we present a subset of features that better explain how easy or difficult an instance is for at least one algorithm. Figures~\ref{fig:corr_cf} and~\ref{fig:goodAlgo} show that instances that have high positive correlation between constraints and objectives are easier to solve, which suggests that the evolutionary trajectory of the search in those instances is not affected by the infeasible area; a search directed by objectives or constraints will probably lead directly to the optimal set of solutions. In addition, Figure~\ref{fig:cpo_upo} suggests that a smaller proportion of $cpo{\_}upo{\_}n$, representing isolation of the non-dominate set or a narrow feasible area, causes difficulty for most algorithms.

Penalty-based algorithms do not have clear footprints in the represented instance space. The group represented by NSGAII, illustrated in Figure~\ref{fig:NSGAII}, finds it easier to solve a problem if the average ratio of the distance between neighbors in the violation space to the distance in the decision space is not low, as shown in Figure~\ref{fig:dist_c}. This suggests the presence of large, neutral areas in the violation landscape. CCMO footprints seem to overlap with the distribution of $dist{\_}f{\_}dist{\_}x{\_}avg{\_}rws$, illustrated in Figure~\ref{fig:dist_f}. The higher this feature is, the more likely CCMO~\ref{fig:CCMO} will succeed. Moreover, CCMO seems to be capable of finding solutions in instances that have a low ratio of solutions in the ideal zone of one objective, as shown in Figure~\ref{fig:piz_ob}, meaning that CCMO has the ability to find isolated optima. Although, this feature has the opposite impact on PPS, as observed in Figure~\ref{fig:PPS}, suggesting that PPS is better suited to find diverse solutions when there are more local optimal solutions, which is also supported by $nhv{\_}avg{\_}rws$ as illustrated in Figure~\ref{fig:nhv_avg}. Furthermore, when there is negative correlation between constraints and objectives, illustrated in Figure~\ref{fig:corr_cf}, PPS is one of the best performing algorithms. We can observe that the instances that have the highest conflict between constraints and objectives, are the instances that ToP was capable to excel at, as shown in Figure~\ref{fig:ToP}.

\subsection{ISA-based Algorithm Selection}\label{subsec:SVM}

\begin{figure}[t]
	\includegraphics[width=0.48\textwidth]{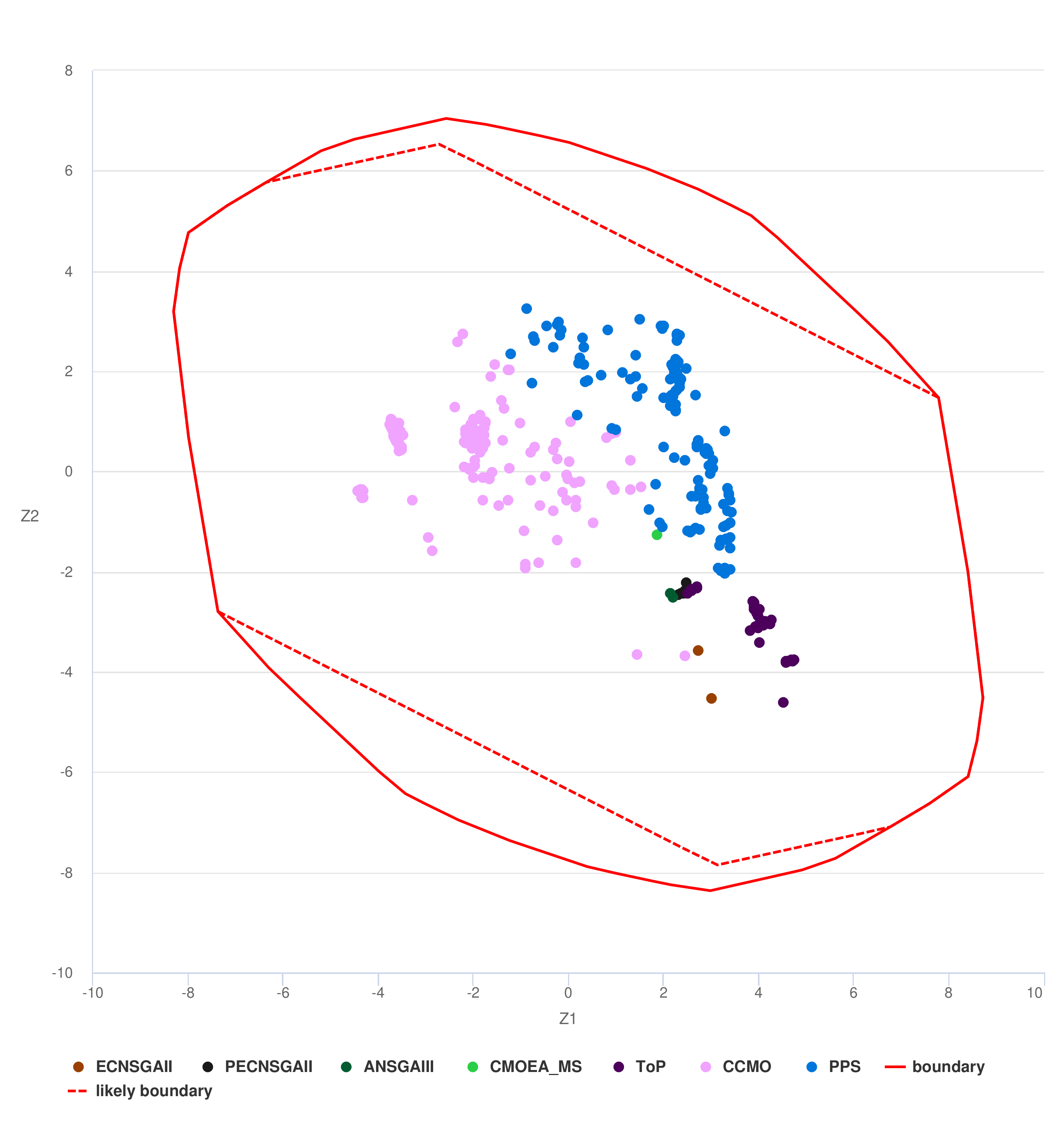}
	\caption{Algorithm recommendations by SVM selection model for the projected instance space.}
	\label{fig:SVM_sel}
\end{figure}

In the final phase of our investigation, we focus on algorithm selection using a support vector machine (SVM). Figure~\ref{fig:SVM_sel} presents the SVM selection results generated by the MATILDA web tools~\cite{MATILDA}. The figure shows that hyper-strategies are more likely to be selected by the SVM model because they surpassed others in larger and clearer regions. The instance space is almost divided between CCMO and PPS, however, there are small area in the bottom-right that predicted to be solved mostly by ToP. However, this area was not easy to solve by almost all the algorithms, as observed in Figure~\ref{fig:afootprints}. Our analysis of the instances in this area found that either there are no feasible solutions, or the HV of the set found by algorithms is approximately zero. Therefore, we conclude that the selection in this area is not accurate.

\begin{figure}[t]
	\includegraphics[width=0.48\textwidth]{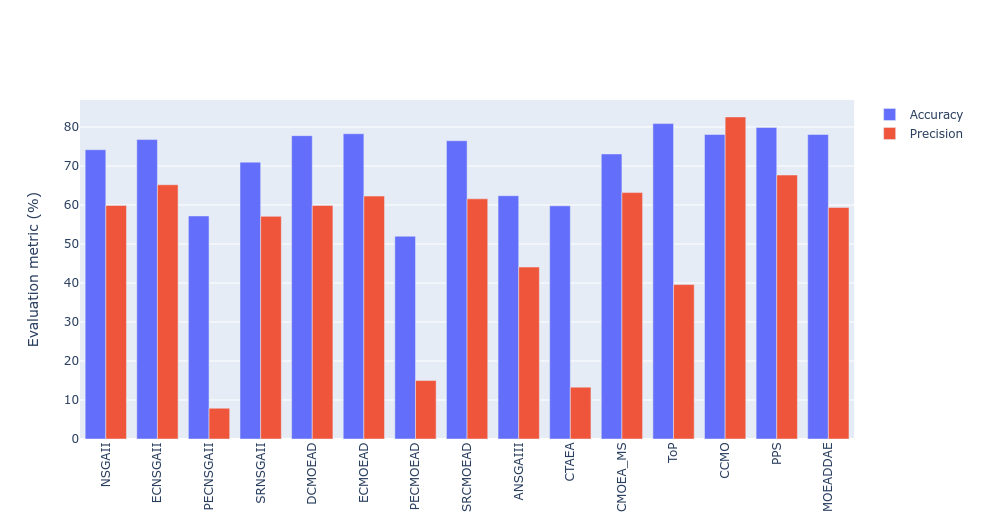}
	\caption{SVM selection model accuracy and precision for each algorithm.}
	\label{fig:SVM_acc}
\end{figure}

In addition, figure~\ref{fig:SVM_acc} describes the accuracy and precision of SVM selection model for each algorithm individually. The results validate the accuracy of the overall model. They illustrate that CCMO and PPS have high accuracy and precision, however, ToP has low precision. In addition, the metrics show that the quality of the algorithm selection model based on the selected features correlate with the clarity of the visualized footprints. For example, PECMOEAD, CTAEA, and ToP have poor SVM metrics values, and they do not have clear footprints on the projected instance space.


\section{Conclusion}\label{sec:conclusion}

We have presented a detailed instance space analysis of CMOPs. Our primary motivation was to systematically evaluate and characterize the conditions where a selected CMOEA was expected to perform well based on the landscape analysis features of COMP instances. Firstly, we have identified COMP features in terms of three landscapes: the multi-objective landscape; the violation landscape and the multi-objective-violation landscape. Secondly, we have collected a large volume of meta-data encapsulating multiple benchmark problem instances and algorithms (including alternative constraint handling techniques). Finally, footprints corresponding to regions of varying algorithm performance were identified. This visual representation provides useful insights, helping to explain CMOPs characteristics and the strengths and weakness of a particular algorithm. In addition, a SVM selection model was then used to select the CMOEA with the best performance in a certain region of the instance space.

Our results show that some CMOEAs, CCMO and PPS in particular, have distinct footprints. CCMO and PPS employ hyper constraint handling techniques, where they use two strategies in two populations/stages. CCMO can effectively converge on isolated optima, whereas PPS generates more diversity when there is a large optimal set. Significantly, our analysis shows that most CMOEAs fail to evolve high quality solutions when there is negative correlation between constraints and objectives. Moreover, CTAEA and other penalty-based algorithms have no clear area of strength, which indicates that the available benchmarks lack examples on which these algorithms would outperform. Therefore, our selector is conservative, as it tends to pick between the two strongest algorithms, i.e., CCMO and PPS, with high precision.

It is widely acknowledged that any benchmark suite of problems should ideally test the efficacy of the optimizer. However, our analysis reveals a lack of diversity in the benchmark suites examined, with many instances sharing similar objective and/or constraint functions. Only a few instances provide a high proportion of constrained Pareto fronts to unconstrained ones, and fewer instances have a highly negative correlation between constraints and objectives. Despite the fact that those two features are challenging to most algorithms.

A wide range of existing and new features have been used in this work, however, they do not result in clear footprints for all algorithms. This, in turn, suggests that there is scope to further explore new features tailored specifically for CMOPs. Additionally, one of the objectives of benchmarking is to create problems that are representative of real-world problems~\cite{Thomas2020}. Thus, we will investigate in future work where real-world problems fall within the instance space.




\ifCLASSOPTIONcaptionsoff
  \newpage
\fi

\bibliographystyle{IEEEtran}
\bibliography{bibliography}

\end{document}